\documentclass[10pt,journal,compsoc]{IEEEtran}

\ifCLASSOPTIONcompsoc
  \usepackage[nocompress]{cite}
\else
  \usepackage{cite}
\fi

\usepackage{graphicx}
\usepackage{textcomp}
\usepackage{xcolor}
\usepackage{amsmath}
\usepackage{url}
\usepackage{bm}
\usepackage{multirow}
\usepackage{algorithm}
\usepackage{makecell}
\usepackage{subcaption}
\usepackage{algpseudocode}
\usepackage{verbatim}
\usepackage{booktabs}
\usepackage{listings}

\newcommand{\RomanNumeralCaps}[1]{\MakeUppercase{\romannumeral #1}}
\usepackage{pifont}

\newcommand{\xmark}{\ding{55}}%
\def\tss{\mathcal{S}_{t}}
\def\sss{\mathcal{S}_{s}}
\definecolor{codegreen}{rgb}{0,0.6,0}
\definecolor{codegray}{rgb}{0.5,0.5,0.5}
\definecolor{codepurple}{rgb}{0.58,0,0.82}
\definecolor{backcolour}{rgb}{0.95,0.95,0.92}
\usepackage{amsmath,amsfonts,bm}

\def\gD{{\mathcal{D}}}

\def\gH{{\mathcal{H}}}

\def\gL{{\mathcal{L}}}

\def\gO{{\mathcal{O}}}

\def\0{\mathbf{0}}
\def\1{\mathbf{1}}

\def\Figref#1{Figure~\ref{#1}}
\def\Secref#1{Section~\ref{#1}}

\def\Tabref#1{Table~\ref{#1}}

\def\NAME{{NATS-Bench}}

\begin{document}

\title{NATS-Bench: Benchmarking NAS Algorithms for Architecture Topology and Size}

\author{Xuanyi Dong, Lu Liu, Katarzyna Musial, Bogdan Gabrys
\IEEEcompsocitemizethanks{
\IEEEcompsocthanksitem Xuanyi Dong, Lu Liu, Katarzyna Musial and Bogdan Gabrys are with School of Computer Science, University of Technology Sydney, NSW, Australia. (e-mail: Xuanyi.Dxy@gmail.com, Lu.Liu.Cs@icloud.com, Katarzyna.Musial-Gabrys@uts.edu.au,  Bogdan.Gabrys@uts.edu.au)\protect
}
}

\markboth{IEEE TRANSACTIONS ON PATTERN ANALYSIS AND MACHINE INTELLIGENCE, {http://dx.doi.org/10.1109/TPAMI.2021.3054824}}{}

\IEEEtitleabstractindextext{
\begin{abstract}
Neural architecture search (NAS) has attracted a lot of attention and has been illustrated to bring tangible benefits in a large number of applications in the past few years. Architecture topology and architecture size have been regarded as two of the most important aspects for the performance of deep learning models and the community has spawned lots of searching algorithms for both of those aspects of the neural architectures. However, the performance gain from these searching algorithms is achieved under different search spaces and training setups. This makes the overall performance of the algorithms incomparable and the improvement from a sub-module of the searching model unclear.
In this paper, we propose NATS-Bench, a unified benchmark on searching for both topology and size, for (almost) any up-to-date NAS algorithm.
NATS-Bench includes the search space of 15,625 neural cell candidates for architecture topology and 32,768 for architecture size on three datasets.
We analyze the validity of our benchmark in terms of various criteria and performance comparison of all candidates in the search space.
We also show the versatility of NATS-Bench by benchmarking 13 recent state-of-the-art NAS algorithms on it.
All logs and diagnostic information trained using the same setup for each candidate are provided.
This facilitates a much larger community of researchers to focus on developing better NAS algorithms in a more comparable and computationally effective environment.
All codes are publicly available at: \url{https://xuanyidong.com/assets/projects/NATS-Bench}.

\end{abstract}

}

\maketitle

\IEEEdisplaynontitleabstractindextext
\IEEEpeerreviewmaketitle

\IEEEraisesectionheading{\section{Introduction}\label{sec:introduction}}

\IEEEPARstart{T}{he} deep learning community is undergoing a transition from hand-designed neural architectures~\cite{he2016deep,krizhevsky2012imagenet,szegedy2015going} to automatically designed neural architectures~\cite{zoph2017NAS,pham2018efficient,real2019regularized,dong2019search,liu2019darts}.
In its early stages, the great success of deep learning was promoted by the introductions of novel neural architectures, such as ResNet~\cite{he2016deep}, Inception~\cite{szegedy2015going}, VGGNet~\cite{simonyan2015very}, and Transformer~\cite{vaswani2017attention}.
However, manually designing one architecture requires human experts to frequently try and evaluate numerous different operation and connection options~\cite{zoph2017NAS}.
In contrast to architectures that are manually designed, those automatically found by neural architecture search (NAS) algorithms require much less human interaction and expert effort.
These NAS-generated architectures have shown promising results in many domains, such as image recognition~\cite{zoph2017NAS,pham2018efficient,real2019regularized} and sequence modeling~\cite{pham2018efficient,dong2019search,liu2019darts}.\looseness=-1

Recently, a variety of NAS algorithms have been increasingly proposed.
While these NAS techniques are methodically designed and show promising improvements, many setups in their algorithms are different.
(1) Different search space is utilized, e.g., range of macro skeletons of the whole architecture~\cite{zoph2018learning,tan2019mnasnet}, a different operation set for the micro cell within the skeleton~\cite{pham2018efficient}, etc.
(2) After a good architecture is selected, various strategies can be employed to train this architecture and report the performance, e.g., different data augmentation~\cite{ghiasi2018dropblock,zhang2018mixup}, different regularization~\cite{zoph2018learning}, different scheduler~\cite{loshchilov2017sgdr}, and different selections of hyperparameters~\cite{liu2018progressive,dong2019one}.
(3) The validation set for testing the performance of the selected architecture is not split in the same way~\cite{liu2019darts,pham2018efficient}.
These discrepancies cause a problem when comparing the performance of various NAS algorithms, making it difficult to conclude their relative contributions.

In response to this challenge, NAS-Bench-101~\cite{ying2019bench} and NAS-HPO-Bench~\cite{klein2019tabular} were proposed.
However, some NAS algorithms cannot be applied \textit{directly} on NAS-Bench-101, and NAS-HPO-Bench only has 144 candidate architectures which may be insufficient to comprehensively evaluate NAS algorithms.
NAS-Bench-1shot1~\cite{arber2020nas1shot1} reuses the NAS-Bench-101 dataset with some modification to analyse the one-shot NAS methods.
The aforementioned works have mainly focused on the architecture topology\footnote{Some works~\cite{shu2020understanding,gu2020dots} use topology to indicate the connectivity pattern of architecture. In this manuscript, the terminology ``architecture topology'' or ``topology'' refers to the connection topology and the associated operation on each connection.}.
However, the architecture size\footnote{Some works~\cite{dong2017more} may use size to indicate the number of parameters of a neural network. In this manuscript, the terminology ``architecture size'' or ``size'' refers to the number of channels in each layer following~\cite{dong2019network}.}, which significantly affects a model's performance, is not considered in the existing benchmarks.\looseness=-1

\begin{figure}[t!]
\begin{center}
\includegraphics[width=\linewidth]{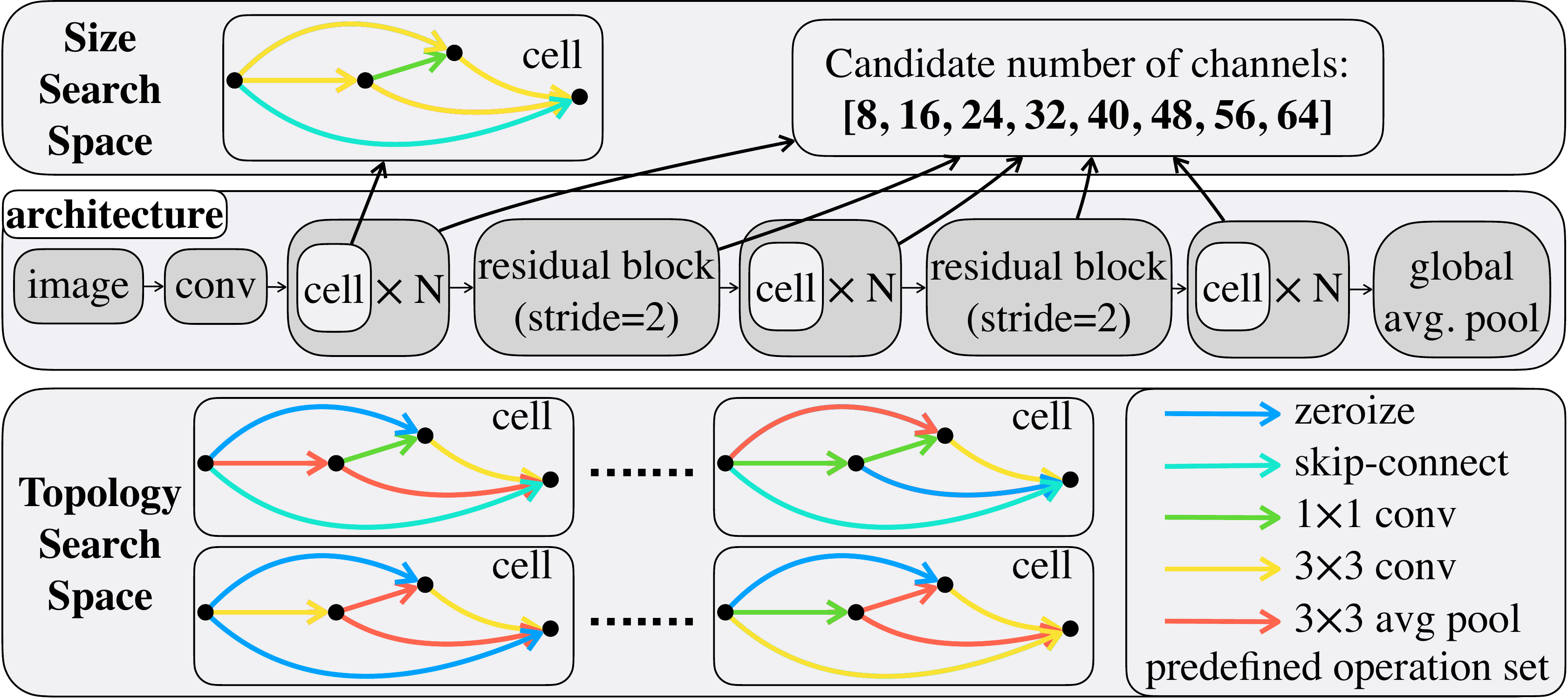}
\end{center}
\caption{
\textbf{Middle}: the macro skeleton of each architecture candidate.
\textbf{Top}: The size search space $\sss$ in {\NAME}.
In $\sss$, each candidate architecture has different configuration for the channel size.
\textbf{Bottom}:  The topology search space $\tss$ in {\NAME}. In $\tss$, each candidate architecture has different cell topology. 
}
\label{fig:search-space}
\end{figure}

To enlarge the scope of these benchmarks and towards better reproducibility of NAS methods,
we propose {\NAME} with (1) a topology search space $\tss$ to be applicable for all NAS methods and (2) a size search space $\sss$ that supplements the lack of analysis for the architecture size.
As shown in \Figref{fig:search-space}, each architecture consists of a predefined skeleton with a stack of the searched cells.
Each cell is represented as a densely-connected directed acyclic graph (DAG) as shown in the bottom section of \Figref{fig:search-space}. 
The node represents the sum of the feature maps and each edge is associated with an operation transforming the feature maps from the source node to the target node.\looseness=-1

In $\tss$, we search for the operation assigned on each edge, and thus its size is related to the number of nodes defined for the DAG and the size of the operation set.
We choose 4 nodes and 5 representative operation candidates for the operation set, which generates a total search space of 15,625 cells/architectures.
In $\sss$, we search for the number of channels in each layer (i.e., convolution, cell, or block). We pre-define 8 candidates for the number of channels, which generates a total search space of $8^{5}=32768$ architectures.
Each architecture in $\tss$ and $\sss$ is trained multiple times on three different datasets.
The training log and performance of each architecture are provided for each run.
The training accuracy/test accuracy/training loss/test loss after every training epoch for each architecture plus the number of parameters and floating point operations (FLOPs) are accessible.\looseness=-1

{\NAME} has shown its value in the field of NAS research.
(1) It provides the \textit{first} benchmark to study the architecture size.
(2) It provides a unified benchmark for most up-to-date NAS algorithms including all cell-based NAS methods.
With {\NAME}, researchers can focus on designing robust searching algorithm while avoiding tedious hyperparameter tuning of the searched architecture.
Thus, {\NAME} provides a relatively fair benchmark for the comparison of different NAS algorithms.
(3) It provides the full training log of each architecture. Unnecessary repetitive training procedure of each selected architecture can be avoided~\cite{liu2018progressive,zoph2017NAS} so that researchers can target on the essence of NAS, i.e., search algorithm.
Another benefit is that the validation time for NAS largely decreases when testing in {\NAME}, which provides a computational power friendly environment for more participation in NAS.
(4) It provides results of each architecture on multiple datasets.
The model transferability can be thoroughly evaluated for most NAS algorithms.
(5) In {\NAME}, we provide systematic analysis of the proposed search space. We also evaluate 13 recent advanced NAS algorithms including reinforcement learning (RL)-based methods, evolutionary strategy (ES)-based methods, differentiable-based methods, etc.
Our empirical analysis can bring some insights to the future designs of NAS algorithms.

\begin{table*}[t!]
\centering
\setlength{\tabcolsep}{7.4pt}
\begin{tabular}{c  c  c  c  c c c c c}
\toprule
                   & \multirow{2}{*}{\makecell{\#Unique\\Architectures}} & \multirow{2}{*}{\#Datasets} & \multirow{2}{*}{Diagnostic Information} & \multirow{2}{*}{Search Space} & \multicolumn{4}{c}{Supported NAS Algorithms} \\
                   & & & & & RL     & ES   & Diff. & HPO\\
\midrule
 NAS-Bench-101            & 423k    & 1  & \xmark & topology & partial & partial & none & most \\
\midrule
 $\tss$ in {\NAME} & 6.5k    & 3  & \multirow{2}{*}{\makecell{fine-grained accuracy\\and loss, parameters, etc}} & topology & all     & all     & all  & most \\

 $\sss$ in {\NAME} & 32.8k   & 3  &   & size   &  all     & all     & most  & most \\
\bottomrule
\end{tabular}
\caption{
We summarize the important characteristics of NAS-Bench-101 and {\NAME}.
Our {\NAME} provides the search space for both architecture topology and architecture size.
Besides, {\NAME} provides train/validation/test performance on three (one for NAS-Bench-101) different datasets so that the generality of NAS algorithms can be evaluated.
It also provides some diagnostic information that may provide insights to design better NAS algorithms.
}
\label{table:compare-bench}
\end{table*}

\section{Related Work}\label{sec:relate-work}

In the past few years, different kinds of search spaces and search algorithms have been proposed. They brought great advancements in many applications of neural network, such as visual perception~\cite{zoph2017neural,tan2020efficientdet,liu2019auto}, language modelling~\cite{pham2018efficient,liu2019darts,dong2019search}, etc.
Despite their success, many researchers have raised concerns about the reproducibility and generalization ability of the NAS algorithms~\cite{lindauer2019best,ying2019bench,dong2020nasbench201,li2019random,li2019random,arber2020nas1shot1}.
It is essentially not clear if the reported improvements have come from hyperparameter settings, re-training pipelines, random seeds, or the improvements of the searching algorithm itself~\cite{lindauer2019best}.
Many researchers devote their effort to solve this problem, and we will introduce them in \Secref{sec:nats-bench-nas-bench} and \Secref{sec:nats-bench-hpo-bench}.\looseness=-1

\subsection{NAS Benchmark}\label{sec:nats-bench-nas-bench}

To the best of our knowledge, NAS-Bench-101~\cite{ying2019bench} is the only existing large-scale architecture dataset.
Similar to {\NAME}, NAS-Bench-101 also transforms the problem of architecture search into the problem of searching neural cells, represented as a DAG.
Differently, NAS-Bench-101 defines operation candidates on the node, whereas we associate operations on the edge as inspired by \cite{liu2019darts,dong2019search,zoph2018learning}.
We summarize characteristics of our {\NAME} and NAS-Bench-101 in \Tabref{table:compare-bench}.
The main highlights of our {\NAME} are as follows.
(1) The search spaces in {\NAME} includes both architecture topology and size, but NAS-Bench-101 only has a topology search space.
(2) {\NAME} is algorithm-agnostic while NAS-Bench-101 without any modification is only applicable to selected algorithms~\cite{yu2020evaluating,arber2020nas1shot1}.
The original complete search space, based on the nodes in NAS-Bench-101, is huge. So, it is exceedingly difficult to efficiently traverse the training of all architectures. To trade off the computational cost and the size of the search space, they constrain the maximum number of edges in the DAG. 
However, it is difficult to incorporate this constraint in all NAS algorithms, such as NAS algorithms based on parameter sharing~\cite{liu2019darts,pham2018efficient}.
Therefore, many NAS algorithms cannot be directly evaluated on NAS-Bench-101.
Our {\NAME} solves this problem by sacrificing the number of nodes and including all possible edges so that our search space is algorithm-agnostic.
(3) We provide architecture information on three (instead of one) datasets and extra diagnostic information, such as architecture computational cost, fine-grained training and evaluation time, etc., which we hope will give inspirations to better and more efficient designs of NAS algorithms.

Despite the existence of NAS-Bench-101, other researchers have also devoted their effort to building a fair comparison and development environments for NAS. 
Zela~et~al.~\cite{arber2020nas1shot1} proposed a general framework for one-shot NAS methods and reused NAS-Bench-101 to benchmark different NAS algorithms.
Yu~et~al.~\cite{yu2020evaluating} designed a novel evaluation framework to evaluate the search phase of NAS algorithms by comparing with a random search.
The aforementioned works have mainly focused on the network topology. However, as other aspects of DNNs, such as network size and optimizer, significantly affect the network's performance, there is a need for an environment and systematic studies covering these areas of NAS.
Unfortunately, until now these aspects have rarely been considered with regard to the problem of reproducibility and generalization ability.

\subsection{HyperParameter Optimization (HPO) Benchmark}\label{sec:nats-bench-hpo-bench}

NAS-HPO-Bench~\cite{klein2019tabular} evaluated 62208 configurations in the joint NAS and hyperparameter space for a simple 2-layer feed-forward network. Since NAS-HPO-Bench has only 144 architectures, it may be insufficient to evaluate different NAS algorithms.
The NAS-HPO-Bench dataset also includes the number of channels in a multi-layer perceptron (MLP).
In contrast, our {\NAME} has a much larger size search space than NAS-HPO-Bench and provides the useful information on deep architecture instead of shallow MLP.

\section{\NAME}\label{sec:nats-info}

Our {\NAME} is algorithm-agnostic. Put simply, it is applicable to almost any up-to-date NAS algorithm. In this section, we will briefly introduce our {\NAME}.
The search space of {\NAME} is inspired by cell-based NAS algorithms (\Secref{sec:nats-bench-arch}).
{\NAME} evaluates each architecture on three different datasets (\Secref{sec:nats-bench-bench}).
All implementation details of {\NAME} are introduced in \Secref{sec:tiny-nas-impl}.
{\NAME} also provides some diagnostic information which can be used for potentially better designs of future NAS algorithms (discussed in \Secref{sec:nas-diagnostic-info}).

\subsection{Architectures in the Search Space}\label{sec:nats-bench-arch}

\textbf{Macro Skeleton}.
Our search space follows the design of its counterpart as used in the recent neural cell-based NAS algorithms~\cite{liu2019darts,zoph2018learning,pham2018efficient}.
As shown in the middle part of \Figref{fig:search-space}, the skeleton is initiated with one 3-by-3 convolution with 16 output channels and a batch normalization layer~\cite{ioffe2015batch}.
The main body of the skeleton includes three stacks of cells, connected by a residual block.
All cells in an architecture has the same topology.
The intermediate residual block is the basic residual block with a stride of 2~\cite{he2016deep}, which serves to down-sample the spatial size and double the channels of an input feature map. The shortcut path in this residual block consists of a 2-by-2 average pooling layer with stride of 2 and a 1-by-1 convolution.
The skeleton ends up with a global average pooling layer to flatten the feature map into a feature vector.
The classification uses a fully connected layer with a softmax layer to transform the feature vector into the final prediction.

\textbf{The Topology Search Space $\tss$.}
The topology search space is inspired by the popular cell-based NAS algorithms~\cite{dong2019search,liu2019darts,zoph2018learning}.
Since all cells in an architecture have the same topology, an architecture candidate in $\tss$ corresponds to a different cell, which is represented as a densely connected DAG.
The densely connected DAG is obtained by assigning a direction from the $i$-th node to the $j$-th node ($i<j$) for each edge in an undirected complete graph.
Each edge in this DAG is associated with an operation transforming the feature map from the source node to the target node.
All possible operations are selected from a predefined operation set, as shown in \Figref{fig:search-space}(bottom-right).
In our {\NAME}, the predefined operation set $\gO$ has $L=5$ representative operations: (1) zeroize, (2) skip connection, (3) 1-by-1 convolution, (4) 3-by-3 convolution, and (5) 3-by-3 average pooling layer.
The convolution in this operation set is an abbreviation of an operation sequence of ReLU, convolution, and batch normalization.
The DAG has $V=4$ nodes, where each node represents the sum of all feature maps transformed through the associated operations of the edges pointing to this node.
We choose $V=4$ to allow the search space to contain basic residual block-like cells, which require 4 nodes. Densely connected DAG does not restrict the searched topology of the cell to be densely connected, since we include zeroize in the operation set, which is an operation of dropping the associated edge.
We do not impose the constraint on the maximum number of edges~\cite{ying2019bench}, and thus $\tss$ is applicable to most NAS algorithms, including all cell-based NAS algorithms.
For each architecture in $\tss$, each cell is stacked $N=5$ times, with the number of output channels set to 16, 32 and 64 for the first, second and third stages, respectively.\looseness-1

\textbf{The Size Search Space $\sss$.}
The size search space is inspired by transformable architecture search methods~\cite{dong2019network,yu2019universally,he2018amc}.
In the size search space, every stack in each architecture is constructed by stacking $N=1$ cell. All cells in every architecture have the same topology, which is the best one in $\tss$ on the CIFAR-100 dataset.
Each architecture candidate in $\sss$ has a different configuration regarding the number of channels in each layer.\footnote{A layer could be the stem 3-by-3 convolutional layer, the cell, or the residual block.}
We build the size search space $\sss$ to include the largest number of channels in $\tss$.
Therefore, the number of channels in each layer is chosen from \{8, 16, 24, 32, 40, 48, 56, 64\}.
Therefore, the size search space $\sss$ has $8^{5}=32768$ architecture candidates.

\begin{table*}[t!]
\centering
\setlength{\tabcolsep}{2.3pt}
\begin{tabular}{c c c c c c c c c c c}
\toprule
      & optimizer &   Nesterov   & learning rate (LR)             & momentum & weight decay & batch size & norm & random flip & random crop & epoch \\
\midrule
value & SGD       & $\checkmark$ & cosine decay LR from 0.1 to 0  & 0.9      & 0.0005       &  256       & $\checkmark$ & p=0.5 & $\checkmark$ & 12   \\
\bottomrule
\end{tabular}
\caption{
The training hyperparameters $\gH^{0}$ for all candidate architectures in $\sss$ and $\tss$.
}
\label{table:train-setting}
\end{table*}

\subsection{Datasets}\label{sec:nats-bench-bench}

We train and evaluate each architecture on CIFAR-10, CIFAR-100~\cite{krizhevsky2009learning}, and ImageNet-16-120~\cite{chrabaszcz2017downsampled}. We choose these three datasets because CIFAR and ImageNet~\cite{russakovsky2015imagenet} are the most popular image classification datasets.

We split each dataset into training, validation and test sets to provide a consistent training and evaluation settings for previous NAS algorithms~\cite{liu2019darts}.
Most NAS methods use the validation set to evaluate architectures after the architecture is optimized on the training set. The validation performance of the architectures serves as the supervision signals to update the searching algorithm.
The test set is to evaluate the performance of each searching algorithm by comparing the indicators (e.g., accuracy, \#parameters, speed) of their selected architectures.
Previous methods use different splitting strategies, which may result in various searching costs and unfair comparisons.
We hope to use the proposed splits to unify the training, validation and test sets for a fairer comparison.\looseness-1

\textbf{CIFAR-10}: It is a standard image classification dataset and consists of 60K 32$\times$32 colour images in 10 classes. The original training set contains 50K images, with 5K images per class. The original test set contains 10K images, with 1K images per class.
Due to the need of validation set, we split all 50K training images in CIFAR-10 into two groups. Each group contains 25K images with 10 classes. We regard the first group as the new training set and the second group as the validation set.

\textbf{CIFAR-100}: This dataset is just like CIFAR-10. It has the same images as CIFAR-10 but categorizes each image into 100 fine-grained classes. The original training set on CIFAR-100 has 50K images, and the original test set has 10K images.
We randomly split the original test set into two groups of equal size --- 5K images per group. One group is regarded as the validation set, and another one is regarded as the new test set.\looseness=-1

\textbf{ImageNet-16-120}: We build ImageNet-16-120 from the down-sampled variant of ImageNet (ImageNet16$\times$16).
As indicated in~\cite{chrabaszcz2017downsampled}, down-sampling images in ImageNet can largely reduce the computation costs for optimal hyperparameters of some classical models while maintaining similar searching results.
\cite{chrabaszcz2017downsampled} down-sampled the original ImageNet to 16$\times$16 pixels to form ImageNet16$\times$16, from which we select all images with label $\in [1,120]$ to construct ImageNet-16-120.
In sum, ImageNet-16-120 contains 151.7K training images, 3K validation images, and 3K test images with 120 classes.\looseness-1

By default, in this paper, ``the training set'', ``the validation set'', ``the test set'' indicate the new training, validation, and test sets, respectively.

\subsection{Architecture Performance}\label{sec:tiny-nas-impl}

\textbf{Training Architectures.}
In order to unify the performance of every architecture, we provide the performance of every architecture in our search space.
In our {\NAME}, we follow previous literature to set up the hyperparameters and training strategies~\cite{zoph2018learning,loshchilov2017sgdr,he2016deep}.
We train each architecture with the same strategy, which is shown in \Tabref{table:train-setting}.
For simplification, we denote all hyperparameters for training a model as a set $\gH$.
We use $\gH^{0}$, $\gH^{1}$, and $\gH^{2}$ to denote the three kinds of hyperparameters that we use.
Specifically, we train each architecture via Nesterov momentum SGD, using the cross-entropy loss. We set the weight decay to 0.0005 and decay the learning rate from 0.1 to 0 with a cosine annealing~\cite{loshchilov2017sgdr}.
We use the same $\gH^{0}$ on different datasets, except for the data augmentation which is slightly different due to the image resolution.
On the CIFAR datasets, we use the random flip with probability of 0.5, the random crop 32$\times$32 patch with 4 pixels padding on each border, and the normalization over RGB channels.
On ImageNet-16-120, we use a similar strategy but with random crop 16$\times$16 patch and 2 pixels padding on each border.
In $\gH^{0}$, we train each architecture by 12 epochs, which can be used in bandit-based algorithms~\cite{falkner2018bohb,li2018hyperband}.
Since 12 epochs are not sufficient to evaluate the relative ranking of different architectures, we train each candidate with more epochs ($\gH^{1}$ and $\gH^{2}$) to obtain a more accurate ranking.
$\gH^{1}$ and $\gH^{2}$ are the same as $\gH^{0}$ but use 200 epochs and 90 epochs, respectively.
In {\NAME}, we apply $\gH^{0}$ and $\gH^{1}$ on the topology search space $\tss$; and we apply $\gH^{0}$ and $\gH^{2}$ on the size search space $\sss$.\looseness=-1

\textbf{Metrics}.
We train each architecture with different random seeds on different datasets.
We evaluate each architecture $\alpha$ after every training epoch.
{\NAME} provides the training, validation, and test loss as well as accuracy.
Users can easily use our API to query the results of each trial of $\alpha$, which has negligible computational costs.
In this way, researchers could significantly speed up their searching algorithm on these datasets and focus solely on the essence of NAS.\looseness-1

\subsection{Diagnostic Information}\label{sec:nas-diagnostic-info}

Validation accuracy is a commonly used supervision signal for NAS.
However, considering the expensive computational costs for evaluating the architecture, the signal is too sparse.
In our {\NAME}, we also provide some additional diagnostic information in a form of extra statistics obtained during training of each architecture.
Collecting these statistics almost involves no extra computation cost but may provide insights for better designs and training strategies of different NAS algorithms, such as platform-aware NAS~\cite{tan2019mnasnet}, accuracy prediction~\cite{baker2018accelerating}, mutation-based NAS~\cite{cai2018efficient,chen2016net2net}, etc.

\textbf{Architecture computational costs:}
{\NAME} provides three computation metrics for each architecture --- the number of parameters, FLOPs, and latency.
Algorithms that focus on searching architectures with computational constraints, such as models on edge devices, can use these metrics directly in their algorithm designs without extra calculations.
We also provide the training time and evaluation time for each architecture.

\textbf{Fine-grained training and evaluation information.}
{\NAME} tracks the changes in loss and accuracy of every architecture after every training epoch.
These fine-grained training and evaluation information often shows the trends related to the architecture performance and could help with identifying some attributes of the model, such as the speed of convergence, the stability, the over-fitting or under-fitting levels, etc.
These attributes may benefit the designs of NAS algorithms.
Besides, some methods learn to predict the final accuracy of an architecture based on the results of a few early training epochs~\cite{baker2018accelerating}.
These algorithms can be trained faster, and the performance of the accuracy prediction can be evaluated using the fine-grained evaluation information.

\textbf{Parameters of the optimized architecture.}
Our {\NAME} releases the trained parameters for each architecture.
This can provide ground truth label for hypernetwork-based NAS methods~\cite{zhang2019graph,brock2018smash}, which learn to generate parameters of architecture.
Other methods mutate an architecture to become another one~\cite{real2019regularized,cai2018efficient}.
With {\NAME}, researchers could directly use the off-the-shelf parameters instead of training them from scratch and analyze how to transfer parameters from one architecture to another.

\subsection{What/Who can Benefit from {\NAME}?}

Our {\NAME} provides a unified NAS library for the community and can benefit NAS algorithms from the perspective of both performance and efficiency.
NAS has been dominated by multi-fidelity based methods~\cite{li2019random,falkner2018bohb,baker2018accelerating,real2019regularized}, which learn to search based on an approximation of the performance of each candidate to accelerate searching.
Running algorithms on our {\NAME} can reduce the approximation to an accurate performance via only querying from the database. This can avoid sub-optimal training because of the inaccurate estimation of the performance as well as accelerate the training into seconds. Meanwhile, with the provision of our diagnostic information, such as latency, algorithms trained with such extra pieces of information can directly fetch them from our codebase with negligible efforts. Meanwhile, the designs of NAS algorithms can also have more diversity with the benefit of the diagnostic information, and more potential designs will be discussed in \Secref{sec:discussion}.

In the NAS community, there has been growing attention to the field of both searching for topology~\cite{liu2019darts,pham2018efficient} and searching for size~\cite{cai2020once,wan2020fbnetv2}. By benchmarking either topology or size on {\NAME}, it may help researchers to understand the effectiveness of these algorithms and give inspirations for ongoing and future research which lies in this intersection.


{\NAME} provides a unified codebase -- a  NAS library -- to make the benchmarking as fair as possible.
In this codebase, we share the code implementation for different algorithms as much as possible.
For example, the super network for weight-sharing methods is reused; the data pipelines for different methods are reused; the interface of training, forwarding, optimizing for different algorithms is kept the same.
We demonstrate, using 13 state-of-the-art NAS algorithms applied on {\NAME}, how the process has been unified through an easy-to-use API.
The implementation difference between DARTS~\cite{liu2019darts} and GDAS~\cite{dong2019search} is only less than 20 lines of code.
Our library reduces the effect caused by the implementation difference when comparing different methods.
It is also easy to implement new NAS algorithms by reusing and extending our library.
More detailed engineering designs can be found in the documentation of our released codes. As this part is beyond the scope of this manuscript, we do not introduce it here.

\section{Analysis of {\NAME}}\label{sec:nats-analysis}

\subsection{An Overview of Architecture Performance}\label{sec:nas-analysis-overview}

\begin{figure*}[t!]
\begin{subfigure}{\linewidth}
  \centering
  \includegraphics[width=\linewidth]{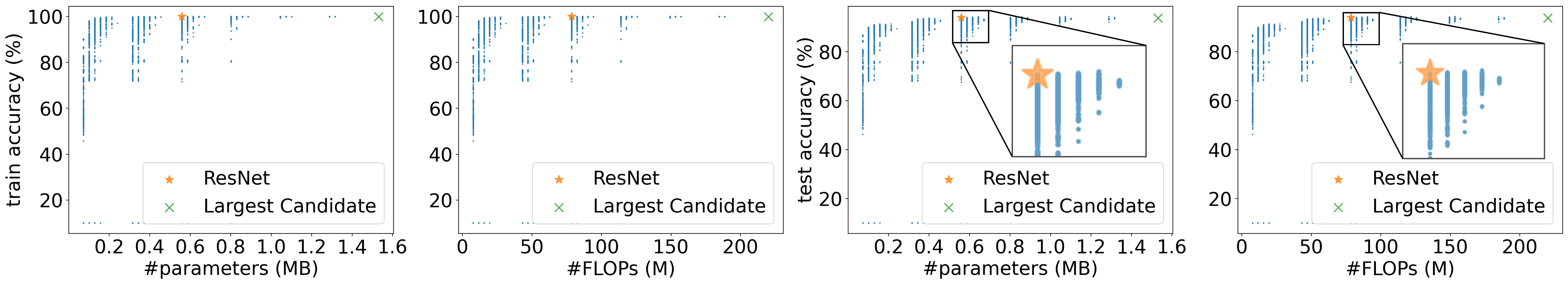}
  \caption{Results of all architecture candidate in the topology search space $\tss$ on the CIFAR-10 dataset.}
  \label{fig:tss-cifar010}
\end{subfigure}
\begin{subfigure}{\linewidth}
  \centering
  \includegraphics[width=\linewidth]{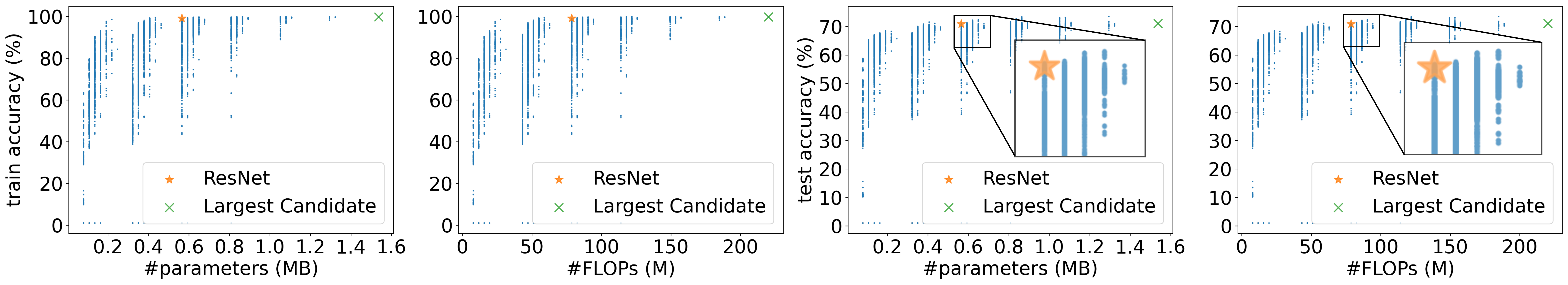}
  \caption{Results of all architecture candidate in the topology search space $\tss$ on the CIFAR-100 dataset.}
  \label{fig:tss-cifar100}
\end{subfigure}
\begin{subfigure}{\linewidth}
  \centering
  \includegraphics[width=\linewidth]{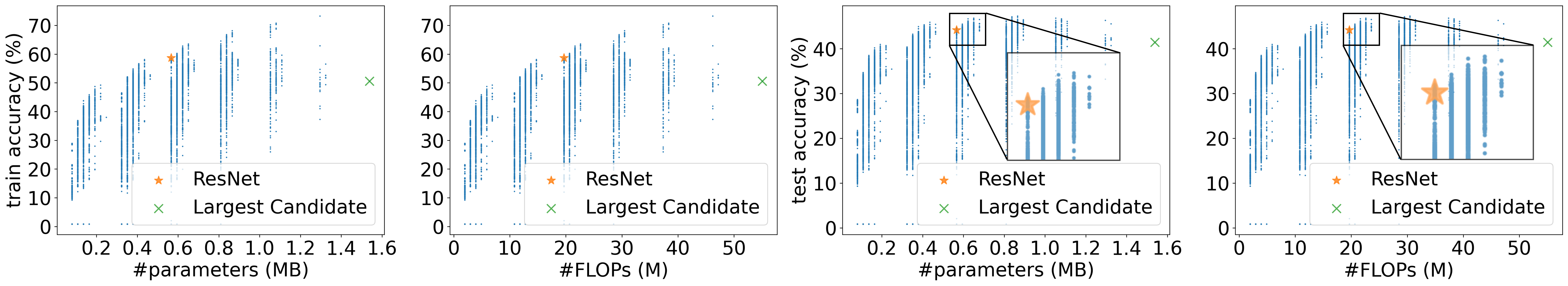}
  \caption{Results of all architecture candidate in the topology search space $\tss$ on the ImageNet-16-120 dataset.}
  \label{fig:tss-imagenet}
\end{subfigure}
\begin{subfigure}{\linewidth}
  \centering
  \includegraphics[width=\linewidth]{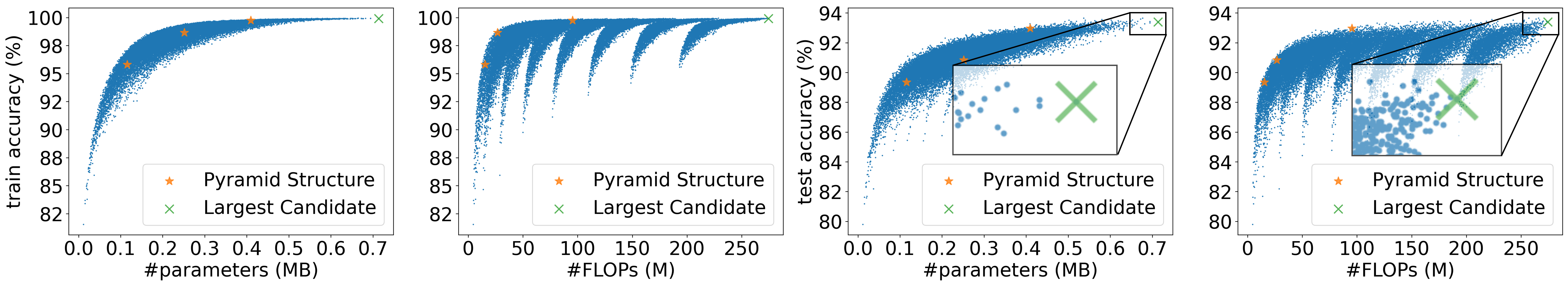}
  \caption{Results of all architecture candidate in the size search space $\sss$ on the CIFAR-10 dataset.}
  \label{fig:sss-cifar010}
\end{subfigure}
\begin{subfigure}{\linewidth}
  \centering
  \includegraphics[width=\linewidth]{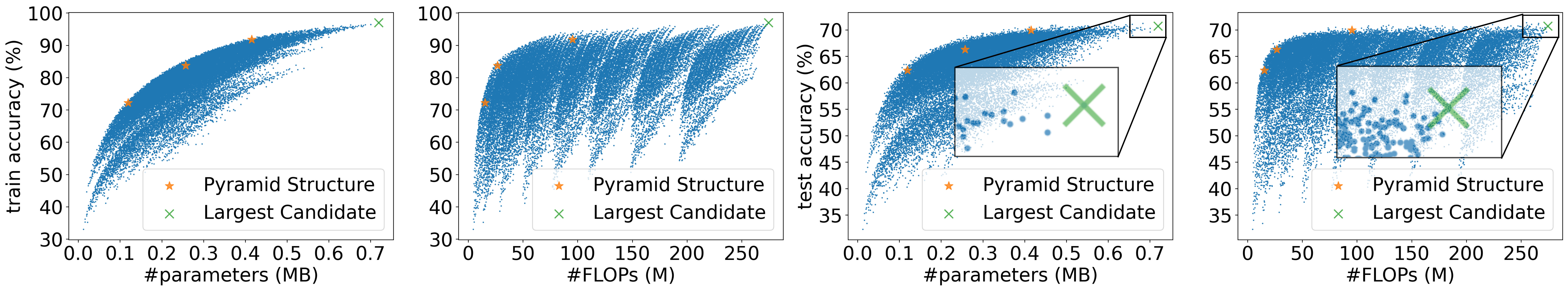}
  \caption{Results of all architecture candidate in the size search space $\sss$ on the CIFAR-100 dataset.}
  \label{fig:sss-cifar100}
\end{subfigure}
\begin{subfigure}{\linewidth}
  \centering
  \includegraphics[width=\linewidth]{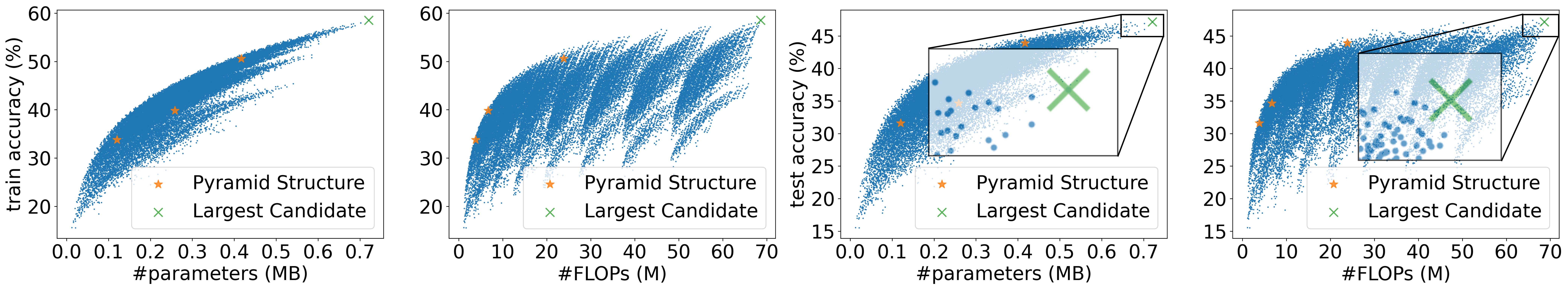}
  \caption{Results of all architecture candidate in the size search space $\sss$ on the ImageNet-16-120 dataset.}
  \label{fig:sss-imagenet}
\end{subfigure}
\caption{
The training and test accuracy vs. the number of parameters and FLOPs for each architecture candidate.
}
\label{fig:all-arch-results}
\end{figure*}

The performance of each architecture in both search spaces $\tss$ and $\sss$ is shown in
\Figref{fig:all-arch-results}.
The training and test accuracy with respect to the number of parameters and number of FLOPs are shown in each column, respectively.
Results show that a different number of parameters or FLOPs will affect the performance of the architectures, which indicates that the choices of operations are essential in NAS.
We also observe that the performance of the architecture can vary even when the number of parameters or FLOPs stays the same.

\begin{figure}[t!]
\begin{subfigure}{0.46\linewidth}
  \centering
  \includegraphics[width=\linewidth]{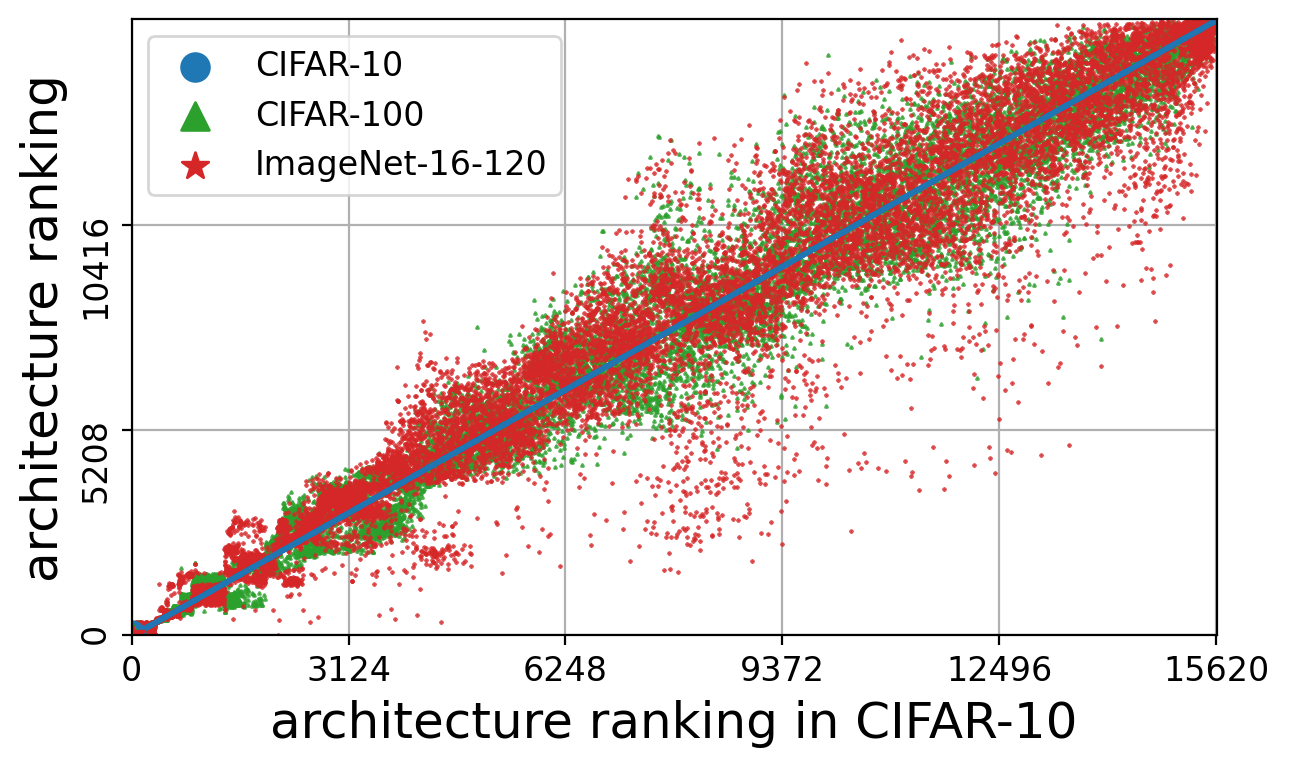}
  \caption{The relative ranking for the topology search space $\tss$.}
  \label{fig:tss-rank-cifar010}
\end{subfigure}
\begin{subfigure}{0.46\linewidth}
  \centering
  \includegraphics[width=\linewidth]{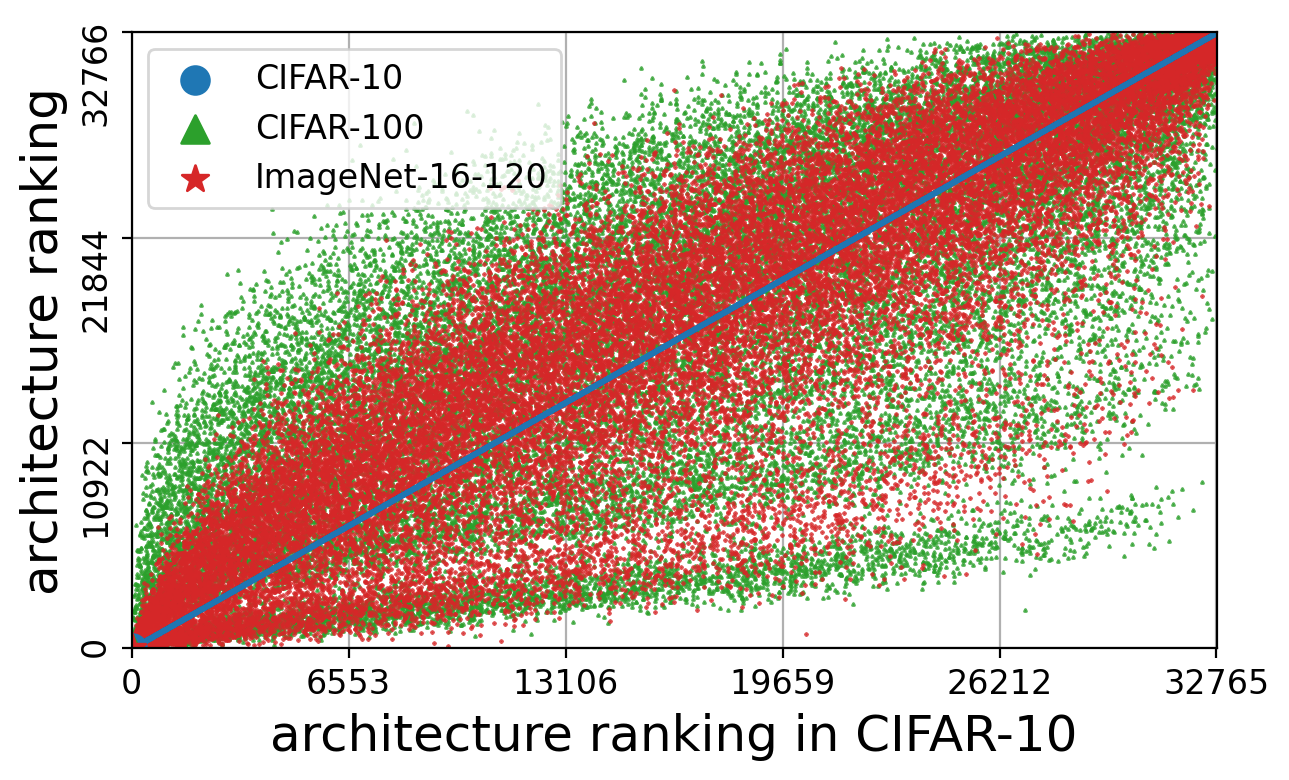}
  \caption{The relative ranking for the size search space $\sss$.}
  \label{fig:sss-rank-cifar100}
\end{subfigure}
\caption{
The ranking of each architecture on three datasets, sorted by the ranking in CIFAR-10.
}
\label{fig:relative-rank}
\end{figure}

These observations indicate the importance of how the operations are connected and how the number of channels is set.
We compare all architectures in $\tss$ and $\sss$ with some classical human-designed architectures (orange star marks in \Figref{fig:all-arch-results}).
(\RomanNumeralCaps{1}) Compared to candidates in $\tss$, ResNet shows competitive performance in three datasets, however, it still has room to improve, i.e., about 2\% compared to the best architecture in CIFAR-100 and ImageNet-16-120, about 1\% compared to the best one with the same amount of parameters in CIFAR-100 and ImageNet-16-120.
(\RomanNumeralCaps{2}) In many vision tasks, the pyramid structure, where the number of channels is gradually increased~\cite{han2017deep}, has shown superior generalization ability.
Inspired by this, we plot three candidates that have a pyramid structure: the number of channels in each layer is 8-16-24-32-40, 8-16-32-48-64, and 32-40-48-56-64.
Regarding the parameters vs. the accuracy, these pyramid candidates in $\sss$ are far from the Pareto optimality.
Regarding the FLOPs vs. the accuracy, they are close to Pareto optimality.

\subsection{Architecture Ranking on Three Datasets}\label{sec:nas-analysis-rank}

The ranking of every architecture in our search space is shown in~\Figref{fig:relative-rank}, where the architectures ranked in CIFAR-10 (x-axis) are shown in relation to their respective ranks in CIFAR-100 and ImageNet-16-120 (y-axis), indicated by green and red markers respectively.
The performance of the architectures in $\tss$ shows a generally consistent ranking over the three datasets with slightly different variance, which serves to test the generality of the searching algorithm.
In contrast, the ranking of architecture candidates in $\sss$ is quite different.
It indicates that the optimal architecture sizes on three datasets are different.

\begin{figure}[t]
\begin{subfigure}{\linewidth}
  \centering
  \includegraphics[width=\linewidth]{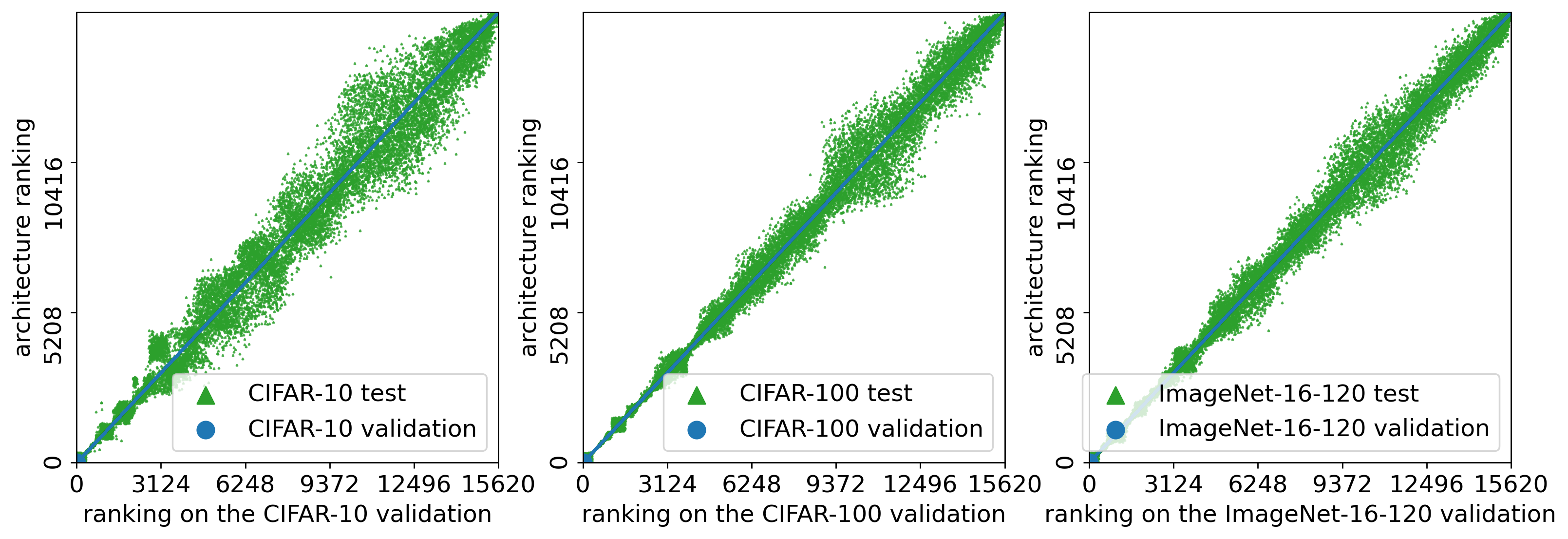}
  \caption{The relative ranking for the topology search space $\tss$.}
  \label{fig:tss-all-sets-rank-cifar010}
\end{subfigure}
\begin{subfigure}{\linewidth}
  \centering
  \includegraphics[width=\linewidth]{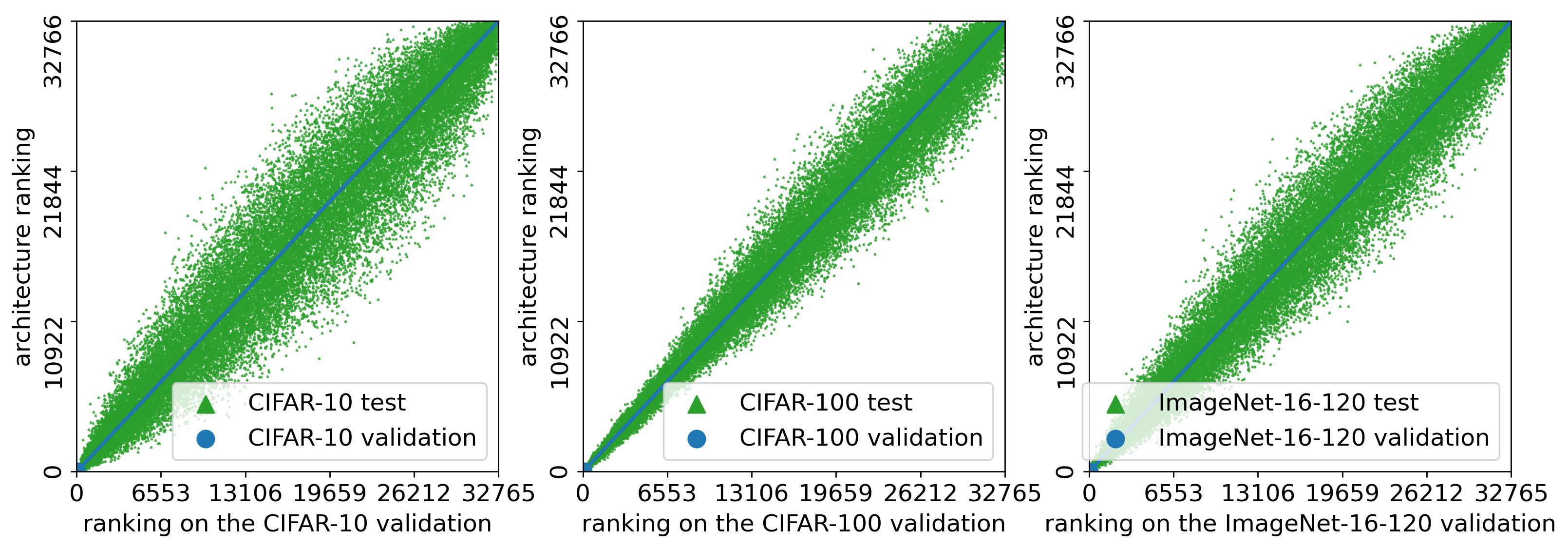}
  \caption{The relative ranking for the size search space $\sss$.}
  \label{fig:sss-all-sets-rank-cifar100}
\end{subfigure}
\caption{
The correlation between the validation accuracy and the test accuracy for all architecture candidates in $\tss$ and $\sss$.
}
\label{fig:relative-rank-sets}
\end{figure}

We compute the validation as well as the test accuracy after training with $\gH^{1}$ and $\gH^{2}$ on $\tss$ and $\sss$, respectively.
\Figref{fig:relative-rank-sets} visualizes their correlation.
It shows the relative ranking obtained from the validation accuracy is similar to that obtained using the test accuracy.
Thus, it guarantees the upper bounds of the NAS algorithms as the brute-force strategy can find an architecture that can almost achieve the highest test accuracy.

\begin{figure}[t!]
\begin{subfigure}{\linewidth}
  \centering
  \includegraphics[width=\linewidth]{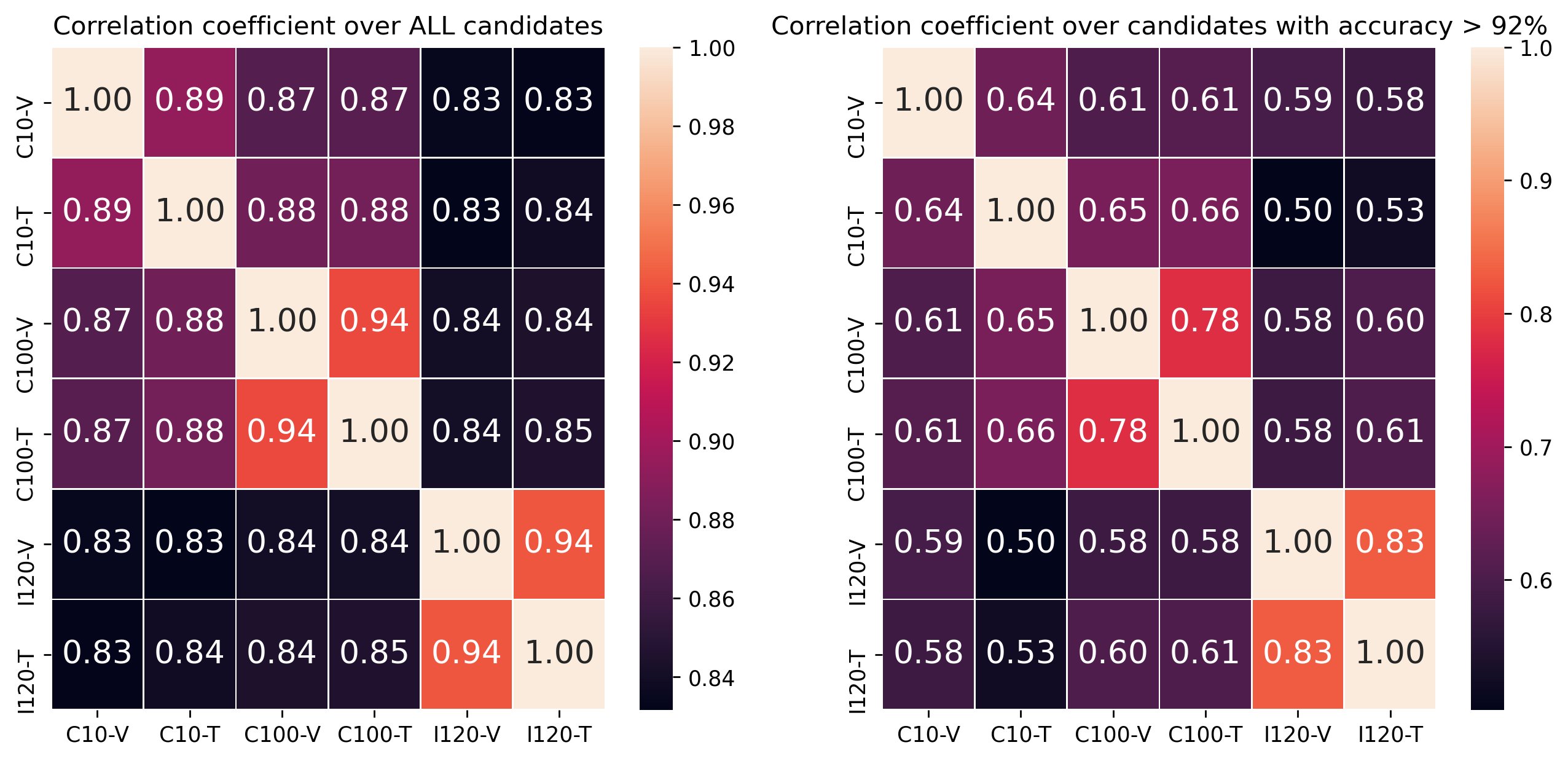}
  \caption{The Kendall rank correlation coefficient for $\tss$.}
  \label{fig:tss-all-sets-all-rank-cifar010}
\end{subfigure}
\begin{subfigure}{\linewidth}
  \centering
  \includegraphics[width=\linewidth]{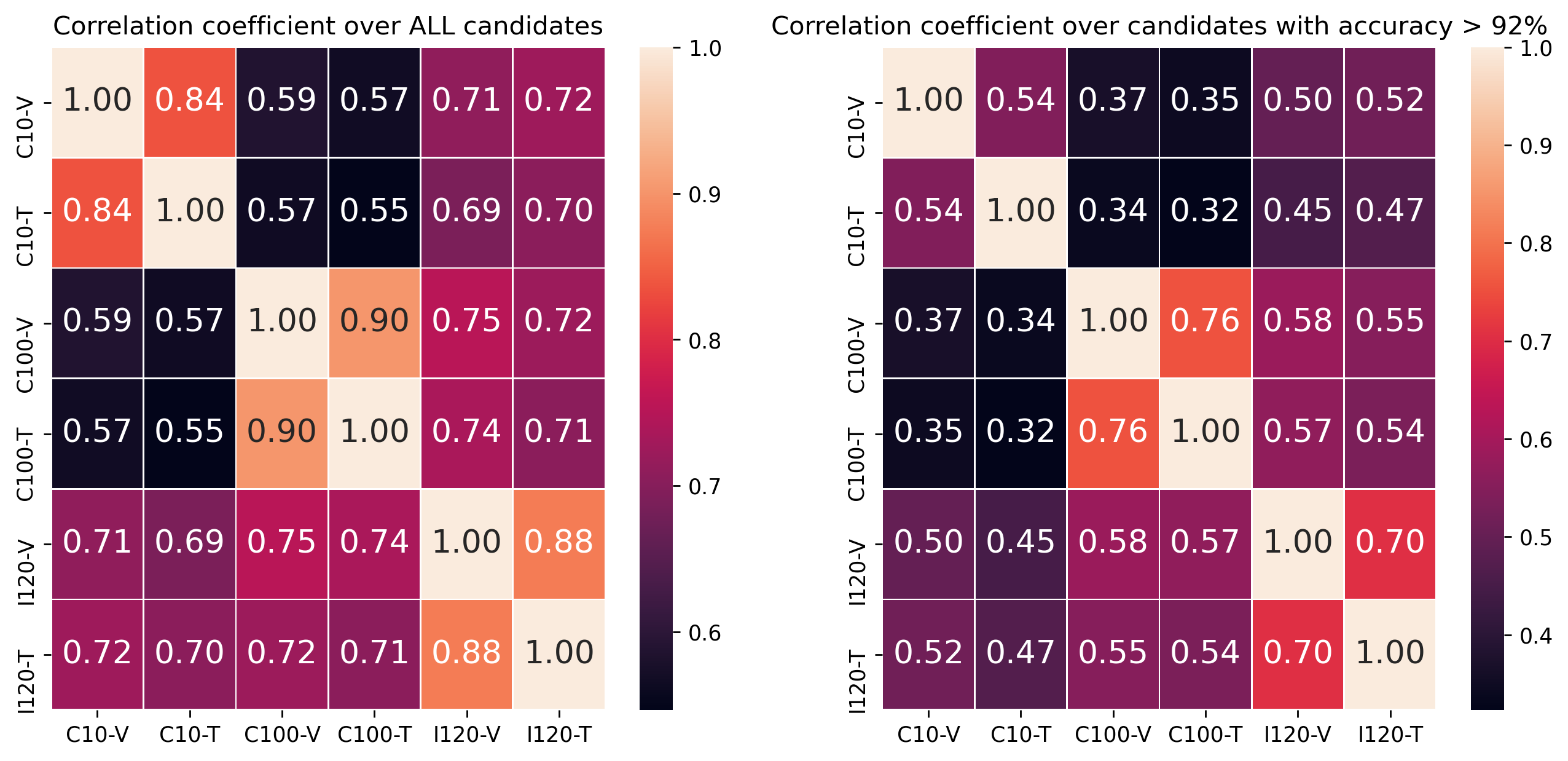}
  \caption{The Kendall rank correlation coefficient for $\sss$.}
  \label{fig:sss-all-sets-all-rank-cifar100}
\end{subfigure}
\caption{
We report the Kendall rank correlation coefficient between the accuracy on 6 sets, i.e., CIFAR-10 validation set (C10-V), CIFAR-10 test set (C10-T), CIFAR-100 validation set (C100-V), CIFAR-100 test set (C100-T), ImageNet-16-120 validation set (I120-V), ImageNet-16-120 test set (I120-T).
}
\label{fig:relative-all-rank-sets}
\end{figure}

We also show the Kendall rank correlation coefficient~\cite{kendall1945treatment} across different datasets in \Figref{fig:relative-all-rank-sets}.
This rank correlation dramatically decreases as we only pick the top performing architecture candidates.
When we directly transfer the best architecture in one dataset to another (i.e. a vanilla strategy), it can not 100\% secure a good performance. This phenomena is a call for better transferable NAS algorithms instead of using the vanilla strategy.

\begin{figure}[t!]
\centering
  \includegraphics[width=\linewidth]{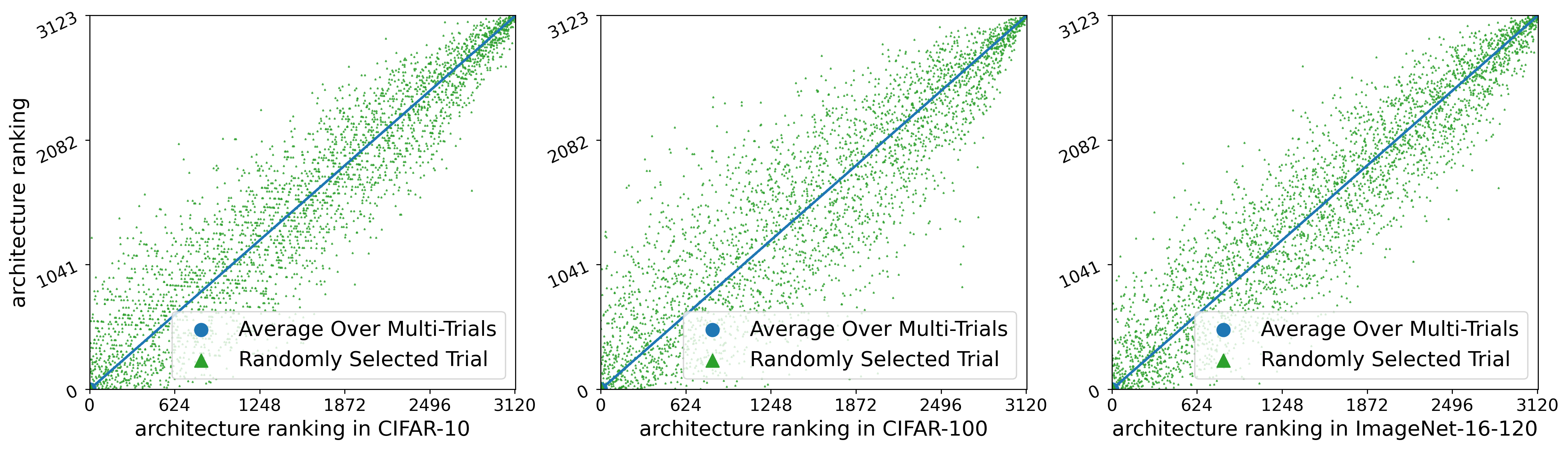}
\caption{
Ranking stability of top 20\% architectures on different datasets over the topology search space $\tss$.
}
\label{fig:tss-rank-stability}
\end{figure}

\textbf{Ranking stability of top architectures.}
The accuracy of an architecture trained in different trials may have a high variance. Such variance of accuracy may affect the architecture rankings.
To investigate such effect, we compare two kinds of rankings:
(1) average the accuracy of multiple trials, and use the averaged accuracy to compute the architecture rankings;
(2) randomly select a trial for each architecture, and use the accuracy of the randomly selected trial to compute the architecture rankings.
We show these two rankings of three datasets in \Figref{fig:tss-rank-stability}.
The Kendall rank correlation coefficient on CIFAR-10, CIFAR-100, and ImageNet-16-120 are about 0.77, 0.70, and 0.77, respectively.

\section{Benchmark}\label{sec:benchmark}

\subsection{Background}\label{sec:bench-back}

NAS aims to find architecture $\alpha$ among the search space $\mathcal{S}$ so that this found $\alpha$ achieves a high performance on the validation set.
This problem can be formulated as a bi-level optimization problem:
\begin{align}\label{eq:nas}
    & \min_{\alpha\in\mathcal{S}} \gL(\alpha, \omega_{\alpha}^{*}, \gD_{val})  \\
    \mathrm{s.t.} \hspace{2mm} & \omega_{\alpha}^{*} = {\arg\min}_{\omega} \gL(\alpha, \omega, \nonumber  \gD_{train}) ,
\end{align}
\noindent where $\gL$ indicates the objective function (e.g., cross-entropy loss).
$\gD_{train}$ and $\gD_{val}$ denote the training data and the validation data, respectively.
In the typical NAS setting, after an architecture $\alpha$ is found, $\alpha$ will be re-trained on $\gD_{train}$ (or $\gD_{train}+\gD_{val}$) and evaluated on the test data $\gD_{test}$ to figure out its real performance.

\subsection{Experimental Setup}\label{sec:bench-setup}

We evaluate \textbf{13} recent, state-of-the-art searching methods on our {\NAME}, which can serve as baselines for future NAS algorithms in our dataset.
Specifically, we evaluate some typical NAS algorithms: (\RomanNumeralCaps{1}) Random Search algorithms, e.g., random search (RANDOM)~\cite{bergstra2012random}, random search with parameter sharing (RSPS)~\cite{li2019random}. (\RomanNumeralCaps{2}) ES methods, e.g., REA~\cite{real2019regularized}.
(\RomanNumeralCaps{3}) RL algorithms, e.g., REINFORCE~\cite{williams1992simple}, ENAS~\cite{pham2018efficient}.
(\RomanNumeralCaps{4}) Differentiable algorithms. e.g., first order DARTS (DARTS 1st)~\cite{liu2019darts}, second order DARTS (DARTS 2nd), GDAS~\cite{dong2019search}, SETN~\cite{dong2019one}, TAS~\cite{dong2019network}, FBNet-V2~\cite{wan2020fbnetv2}, TuNAS~\cite{bender2020can}.
(\RomanNumeralCaps{5}) HPO methods, e.g., BOHB~\cite{falkner2018bohb}.

Among them, RANDOM, REA, REINFORCE, and BOHB are multi-trial based methods. They can be used to search on both $\tss$ and $\sss$ search spaces. Especially, using our API, we can accelerate them to be executed in seconds as shown in \Tabref{table:NAS-Benefits}.

Other methods are weight-sharing based methods, in which the evaluation procedure can be accelerated by using our API.
Notably, DARTS, GDAS, SETN are specifically designed for the topology search space $\tss$.
The search strategies for \#channels in TAS, FBNet-V2, and TuNAS can be used on the size search space $\sss$.

\begin{table}[t!]
\centering
\setlength{\tabcolsep}{1.1pt}
\begin{tabular}{ c | c }
\toprule
Accelerate the search & RANDOM, REINFORCE, REA, and BOHB \\
Accelerate the evaluation & all NAS methods\\
\bottomrule
\end{tabular}
\caption{
The utility of our {\NAME} for different NAS algorithms.
We show whether a NAS algorithm can use our {\NAME} to accelerate the searching and evaluation procedure.
}
\label{table:NAS-Benefits}
\end{table}

\begin{figure*}[t!]
\begin{subfigure}{\linewidth}
  \centering
  \includegraphics[width=\linewidth]{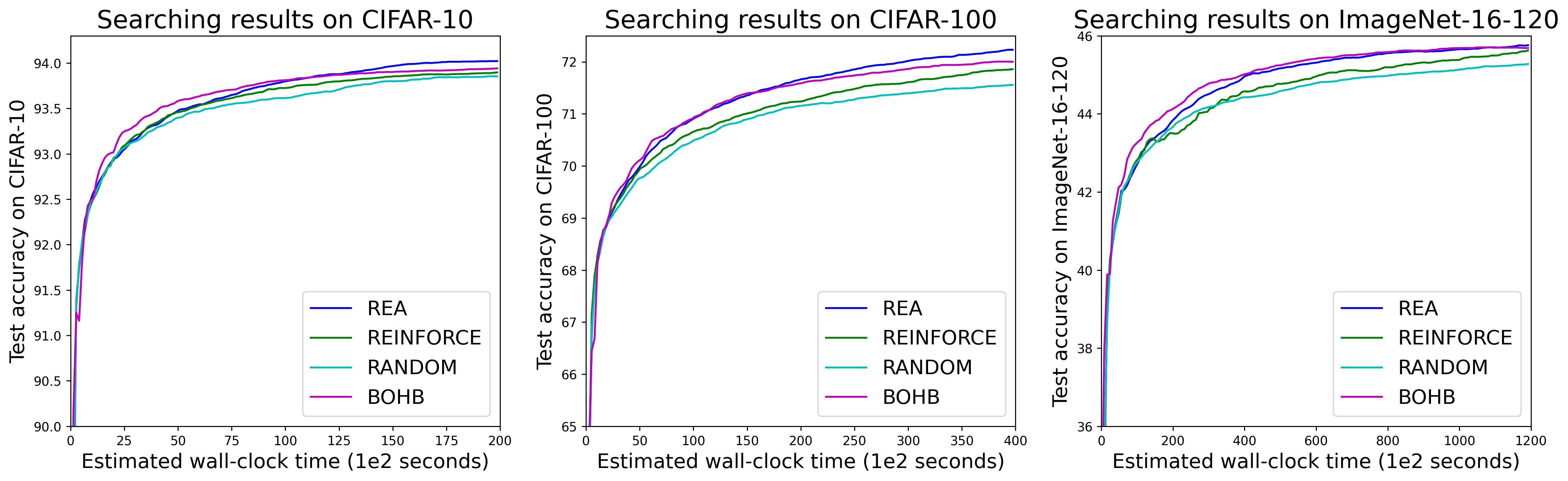}
  \caption{Results of NAS algorithms without weight sharing in the topology search space $\tss$.}
  \label{fig:nas-alg-tss}
\end{subfigure}
\begin{subfigure}{\linewidth}
  \centering
  \includegraphics[width=\linewidth]{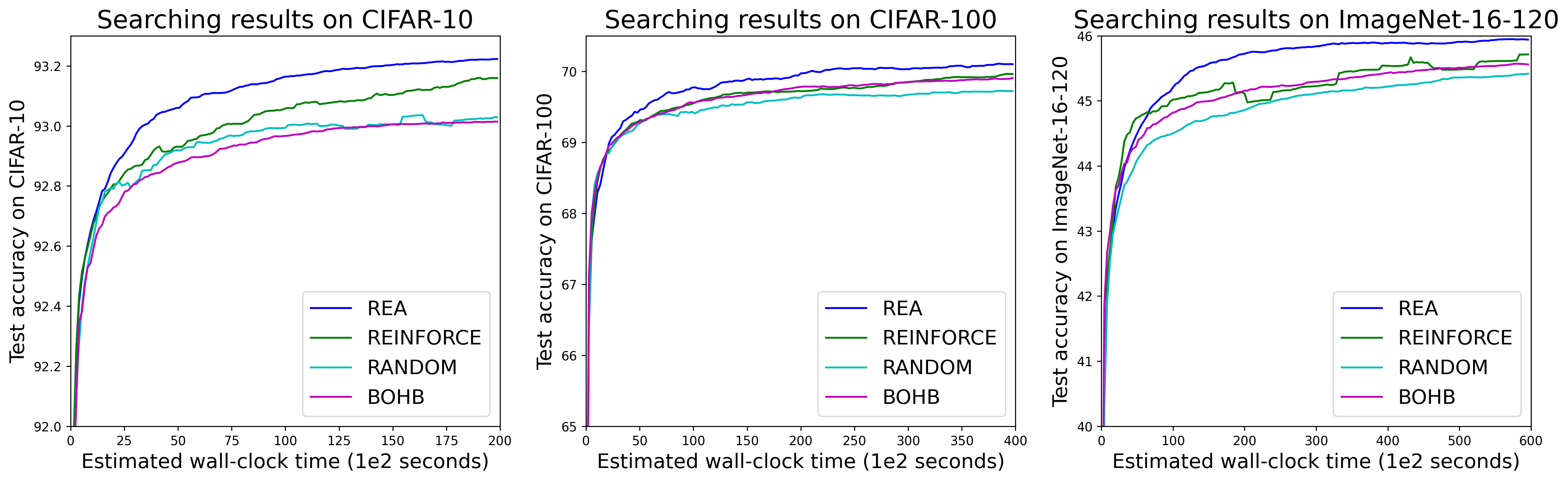}
  \caption{Results of NAS algorithms without weight sharing in the size search space $\sss$.}
  \label{fig:nas-alg-sss}
\end{subfigure}
\caption{
The test accuracy of the searched architecture candidate over time.
We run different searching algorithms 500 times on three datasets.
We plot the test accuracy of their searched model at each timestamp for the corresponding dataset.
This test accuracy is evaluated after fully training the model on the corresponding dataset and averaged over 500 runs.
We report more details including the accurate numbers and variance in \Tabref{table:benchmarking}.
}
\label{fig:fed-nas-alg-results}
\end{figure*}

\subsection{Experimental Results}\label{sec:bench-exp-res}

\subsubsection{Multi-trial based Methods}\label{sec:bench-exp-res-mul}

We follow the suggested hyperparameters in their original papers to run each method on our topology search space $\tss$ and size search space $\sss$.
We run each experiment 500 times on three datasets.
For the CIFAR-10 dataset, we set up a maximum time budget of 2e4 seconds.
As the training time for a single model is much larger on other datasets than that on CIFAR-10, we increase this time budget for other datasets accordingly.
Every 100 seconds, each method can let us know the current searched architecture candidate.
We use the hyperparameters $\gH^{0}$ (12 epochs) to obtain a validation accuracy for each trial.
This validation accuracy serves as the supervision/feedback signal for these multi-trial based methods.
For BOHB, given its current budget for a trial, it can early stop before fully training the model using 12 epochs.
We show the averaged accuracy of this searched architecture candidate over 500 runs in \Figref{fig:fed-nas-alg-results} and \Tabref{table:benchmarking}.
Each sub-figure in \Figref{fig:fed-nas-alg-results} corresponds to one dataset and a search space.
For example, in the middle of \Figref{fig:nas-alg-sss}, we search on CIFAR-100 and show the test accuracy of the automatically discovered architecture on CIFAR-100.

\textbf{Observations on the topology search space $\tss$}.
(1) On CIFAR-10, most methods have similar performance.
(2) On CIFAR-100, before 2e4 seconds, REA is similar to BOHB, and they outperforms REINFORCE and RANDOM; at 4e4 seconds, REA $\geq$ BOHB $\geq$ REINFORCE $\geq$ RANDOM.
(3) On ImageNet-16-120, BOHB converges faster than the other methods. It may be caused by the dynamic budget mechanism for each trial in BOHB, which allows to traverse more architecture candidates.

\textbf{Observations on the size search space $\sss$}.
(1) REA significantly outperforms the other methods on all datasets in the size search space $\sss$.
(2) On CIFAR-10, REINFORCE is better than BOHB and RANDOM.
(3) On CIFAR-100 and ImageNet-16-120, the results of BOHB and REINFORCE are similar, and RANDOM is the worst one.
(4) As the searching time goes, the searched architecture by REA becomes closer and closer to the best one, while the other methods need much more time to catch up with REA.
(5) \Figref{fig:all-arch-results} implies a simple prior for $\sss$: without the constraint of model cost, the larger model tends to have higher accuracy. By visualising the searched architecture, REA can quickly fit this prior while the other methods do not.

Given the flexibility and robustness of REA, we would recommend choosing REA as a searching algorithm if the computational resources are sufficient.

\begin{figure*}[t!]
\begin{subfigure}{\linewidth}
  \centering
  \includegraphics[width=\linewidth]{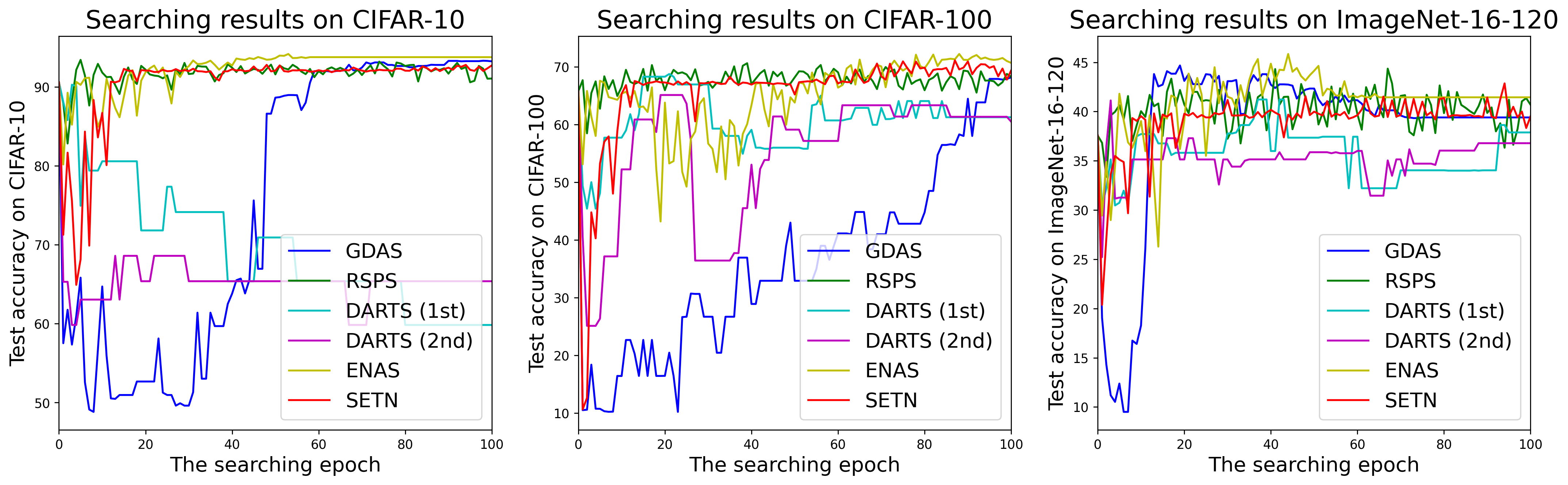}
  \caption{Results of weight-sharing based methods in the topology search space $\tss$.}
  \label{fig:ws-nas-alg-tss}
\end{subfigure}
\begin{subfigure}{\linewidth}
  \centering
  \includegraphics[width=\linewidth]{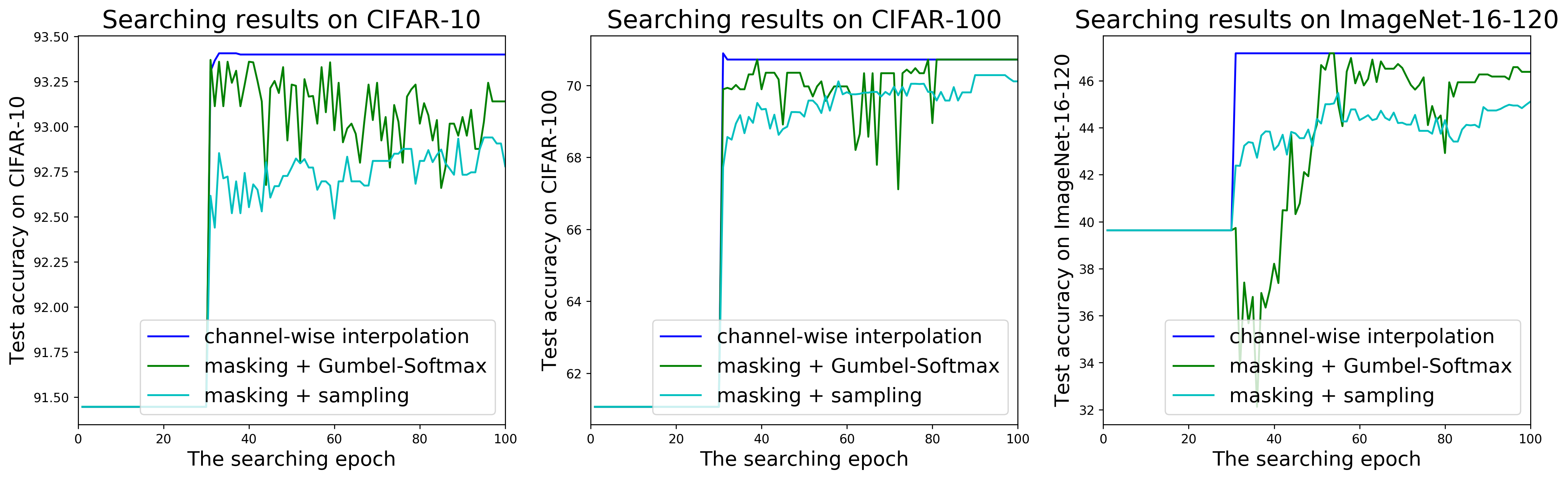}
  \caption{
  Results of weight-sharing based methods in the size search space $\sss$. We do not add any \#FLOPs or \#parameters constraint for these searching methods. Following~\cite{wan2020fbnetv2,bender2020can}, at the first 30\% search phase, we warmup the shared weights with the same strategy as~\cite{bender2020can}.
  }
  \label{fig:ws-nas-alg-sss-warm}
\end{subfigure}
\caption{
The test accuracy of the searched architecture candidate after each search epoch.
We run different searching algorithms three times on three datasets.
We plot the test accuracy of their searched model after each search epoch for the corresponding dataset.
This test accuracy is evaluated after fully training the model on the corresponding dataset and averaged over three runs.
We report more details including the accurate numbers and variance in \Tabref{table:benchmarking}.
}
\label{fig:com-ws-nas-alg-results}
\end{figure*}

\subsubsection{Weight-sharing based Methods}\label{sec:bench-exp-res-sharew}

To compare weight-sharing based methods as fairly as possible, we keep the same hyperparameters concerned with the optimising of the shared weights for different methods.
For other hyperparameters, e.g., hyperparameters for optimising the controller in ENAS or hyperparameters for optimising the architectural parameters in DARTS/GDAS, we use the same values as introduced in their original papers by default.
For the NAS algorithms for the size search space, we follow~\cite{bender2020can} to warmup the one-shot model at the first 30\% search phase.
In this way, we can focus on evaluating the core and unique modules in each searching algorithm.
We setup the total number of epochs to 100 for search, and compare results of their searched architecture candidates after each search epoch.
We run each experiment three times and report the average results in \Figref{fig:ws-nas-alg-tss}, \Figref{fig:ws-nas-alg-sss-warm}, and \Tabref{table:benchmarking}.

\textbf{Observations on the topology search space $\tss$}.
(1) On CIFAR-10, DARTS (1st) and DARTS (2nd) quickly converge to find the architecture having many skip connections, which performs poorly. However, on CIFAR-100 and ImageNet-16-120, they perform relatively well.
This is because the significantly increased searching data on CIFAR-100 and ImageNet-16-120 over CIFAR-10 alleviate the problem of incorrect gradient estimation in bi-level optimization.
(2) RSPS, ENAS, and SETN converge quickly and are robust on three datasets. During their searching procedure, they will randomly sample some architecture candidates, evaluate them using the shared weights, and select the candidate with the highest validation accuracy. Such strategy is more robust than using the $\arg\max$ over the learned architecture parameters in~\cite{liu2019darts,dong2019search}.
(3) The searched architecture of GDAS slowly converges to the similar one as ENAS and SETN.

Some observations on $\tss$ are different from those in our preliminary version. It is because some hyperparameters changed following either suggestions from the authors or better strategies found in our experiments.
Especially, we would like to highlight some useful strategies for weight-sharing based methods:
(1) always use batch statistics for the batch normalization layer.
(2) use the same configuration of the standalone for the one-shot model, such as number of layers and batch size.
(3) during the evaluation procedure of RSPS, ENAS, and SETN, the average accuracy for a large batch of validation data is sufficient to approximate the average accuracy on the whole validation set. In our experiments, we use the batch size of 512 for evaluation.

\textbf{Observations on the size search space $\sss$}.
We abstract the three kinds of strategies to search for \#channels:
\begin{itemize}
    \item Using channel-wise interpolation to explicitly compare two different \#channels~\cite{dong2019network}.
    \item Using the masking mechanism to represent different candidate \#channels and optimize its distribution via Gumbel-Softmax~\cite{wan2020fbnetv2}.
    \item Using the masking mechanism to represent different candidate \#channels and optimize its distribution via REINFORCE~\cite{bender2020can}.
\end{itemize}
We indicate these strategies as ``channel-wise interpolation'', ``masking + Gumbel-Softmax'', and ``masking + sampling'' in \Figref{fig:ws-nas-alg-sss-warm}.
``channel-wise interpolation'' can quickly find much better model than masking-based strategies.
It might because that the interpolation strategy allows us to implicitly evaluate and compare two candidate \#channels in each layer during each search step. In contrast, the masking strategies can only evaluate one candidate during each search step.

\begin{table*}[!t]
\centering
\setlength{\tabcolsep}{4pt}
\begin{tabular}{ c | c | c | c | c | c | c | c | c }
\toprule
\multirow{2}{*}{Search Space} & \multicolumn{2}{|c}{Methods} & \multicolumn{2}{|c}{CIFAR-10} & \multicolumn{2}{|c}{CIFAR-100} & \multicolumn{2}{|c}{ImageNet-16-120} \\
\cmidrule{2-9}
          &  Type & Name  & validation     & test           &   validation   &   test    & validation & test  \\
\midrule
\multirow{12}{*}{\makecell{Topology\\Search\\Space $\tss$}}
 & \multirow{4}{*}{\makecell{Multi-trial}}
   & REA       & 91.25$\pm$0.31 & 94.02$\pm$0.31 & 72.28$\pm$0.95 & 72.23$\pm$0.84 & 45.71$\pm$0.77 & 45.77$\pm$0.80
\\
 & & REINFORCE & 91.12$\pm$0.25 & 93.90$\pm$0.26 & 71.80$\pm$0.94 & 71.86$\pm$0.89 & 45.37$\pm$0.74 & 45.64$\pm$0.78 \\
 & & RANDOM    & 91.07$\pm$0.26 & 93.86$\pm$0.23 & 71.46$\pm$0.97 & 71.55$\pm$0.97 & 45.03$\pm$0.91 & 45.28$\pm$0.97 \\
 & & BOHB      & 91.17$\pm$0.27 & 93.94$\pm$0.28 & 72.04$\pm$0.93 & 72.00$\pm$0.86 & 45.55$\pm$0.79 & 45.70$\pm$0.86 \\\cmidrule{2-9}
 & \multirow{6}{*}{\makecell{Weight\\Sharing}}
   & RSPS     & 87.60$\pm$0.61  & 91.05$\pm$0.66 & 68.27$\pm$0.72 & 68.26$\pm$0.96 & 39.73$\pm$0.34 & 40.69$\pm$0.36\\
 & & DARTS (1st) & 49.27$\pm$13.44 & 59.84$\pm$7.84 & 61.08$\pm$4.37 & 61.26$\pm$4.43 & 38.07$\pm$2.90 & 37.88$\pm$2.91 \\
 & & DARTS (2nd) & 58.78$\pm$13.44 & 65.38$\pm$7.84 & 59.48$\pm$5.13 & 60.49$\pm$4.95 & 37.56$\pm$7.10 & 36.79$\pm$7.59 \\
 & & GDAS     & 89.68$\pm$0.72  & 93.23$\pm$0.58 & 68.35$\pm$2.71 & 68.17$\pm$2.50 & 39.55$\pm$0.00 & 39.40$\pm$0.00 \\
 & & SETN     & 90.00$\pm$0.97  & 92.72$\pm$0.73 & 69.19$\pm$1.42 & 69.36$\pm$1.72 & 39.77$\pm$0.33 & 39.51$\pm$0.33 \\
 & & ENAS     & 90.20$\pm$0.00  & 93.76$\pm$0.00 & 70.21$\pm$0.71 & 70.67$\pm$0.62 & 40.78$\pm$0.00 & 41.44$\pm$0.00 \\\cmidrule{2-9}
 & \multicolumn{2}{|c|}{\textit{ResNet}} & 90.86 & 93.91 & 70.50 & 70.89 & 44.10 & 44.23 \\
 & \multicolumn{2}{|c|}{\textbf{Optimal}} & 91.61 & 94.37 (94.37) & 73.49 & 73.51 (73.51) & 46.73 & 46.20 (47.31) \\
\midrule
\midrule
\multirow{9}{*}{\makecell{Size\\Search\\Space $\sss$}}
 & \multirow{4}{*}{\makecell{Multi-trial}}
   & REA       & 90.37$\pm$0.20 & 93.22$\pm$0.16 & 70.23$\pm$0.50 & 70.11$\pm$0.61 & 45.30$\pm$0.69 & 45.94$\pm$0.92
\\
 & & REINFORCE & 90.25$\pm$0.23 & 93.16$\pm$0.21 & 69.84$\pm$0.59 & 69.96$\pm$0.57 & 45.06$\pm$0.77 & 45.71$\pm$0.93 \\
 & & RANDOM    & 90.10$\pm$0.26 & 93.03$\pm$0.25 & 69.57$\pm$0.57 & 69.72$\pm$0.61 & 45.01$\pm$0.74 & 45.42$\pm$0.86 \\
 & & BOHB      & 90.07$\pm$0.28 & 93.01$\pm$0.24 & 69.75$\pm$0.60 & 69.90$\pm$0.60 & 45.11$\pm$0.69 & 45.56$\pm$0.81 \\\cmidrule{2-9}
 & \multirow{3}{*}{\makecell{Weight\\Sharing}}
   & channel-wise interpolation & 90.71$\pm$0.00 & 93.40$\pm$0.00 & 70.30$\pm$0.00 & 70.72$\pm$0.00 & 44.73$\pm$0.00 & 47.17$\pm$0.00
 \\
 & & masking + Gumbel-Softmax   & 90.41$\pm$0.10 & 93.14$\pm$0.13 & 70.30$\pm$0.00 & 70.72$\pm$0.00 & 45.71$\pm$0.39 & 46.38$\pm$0.27 \\
 & & masking + sampling         & 89.73$\pm$0.37 & 92.78$\pm$0.30 & 69.67$\pm$0.22 & 70.11$\pm$0.33 & 44.70$\pm$0.60 & 45.11$\pm$0.76 \\\cmidrule{2-9}
 & \multicolumn{2}{|c|}{\textit{Largest Candidate}} & 90.71 & 93.40 & 70.30 & 70.72 & 44.73 & 47.17 \\
 & \multicolumn{2}{|c|}{\textbf{Optimal}} & 90.71 & 93.40 (93.65) & 70.92 & 70.12 (71.34) & 46.73 & 45.10 (47.40) \\
\bottomrule
\end{tabular}
\caption{
We evaluate \textit{13} different searching algorithms in our {\NAME}.
We use these algorithms to search for the architectures on different datasets and report \textit{the accuracy of their discovered architectures} on the corresponding dataset.
For multi-trial based methods, we run each algorithm 500 times and report the mean$\pm$variance.
For weight-sharing based methods, we run each algorithm 3 times and report the mean$\pm$variance.
``ResNet'' indicates the candidate in topology search space with the same cell structure as the residual network~\cite{he2016deep}.
``Largest Candidate'' indicates the candidate in size search space with each \#channels of 64.
For ``Optimal'', we report the performance of the candidate with the highest validation accuracy on each dataset, and we also include the highest test accuracy in the parentheses.\\
Notably, compared to the old version of this table in~\cite{dong2020nasbench201}, we include the searching results on three (instead of one) datasets with better hyperparameters.
}
\label{table:benchmarking}
\end{table*}


Since the original hyperparameters of \cite{dong2019network,wan2020fbnetv2,bender2020can} are designed for ImageNet and joint searching of filters and operations, they might be sub-optimal for the settings in {\NAME}. In addition, we re-implement these algorithms based on our codebase. They may have some differences compared to the original implementation due to the different search spaces, libraries, etc.
Therefore, it is under investigation of whether our empirical observations can generalize to other scenarios or not.

\subsubsection{Weight-sharing vs. Multi-trial based Methods}\label{sec:bench-exp-res-vs}

The weight-sharing based methods and multi-trial based methods have their unique advantages and disadvantages.
Multi-trial based methods can theoretically find the best architecture as long as the proxy task is accurate, and the number of trials is large enough.
However, their prohibitive computational cost has motivated researchers to design efficient weight-sharing based algorithms.
However, sharing weights sacrifices the accuracy of each architecture candidate. As the search space increases, the shared weights are usually not able to distinguish the performance of different candidates.\looseness-1

\textbf{Clarification.} We have tried our best to implement each method using their reported best experimental set ups. However, please be aware that some algorithms might still result in sub-optimal performance since their hyperparameters might not be optimal for our {\NAME}.
We empirically found that some NAS algorithms are sensitive to some hyperparameters, and we have tried to compare them in as fair a way as possible.
If researchers can provide better results with different hyperparameters, we are happy to update the benchmarks according to the new experimental results. We also welcome more NAS algorithms to be tested on our dataset and would be happy to include them accordingly.

\section{Discussion}\label{sec:discussion}

\textbf{How to avoid over-fitting on {\NAME}?}
Our {\NAME} provides a benchmark for NAS algorithms, aiming to provide a fair and computationally cost-friendly environment to the NAS community.
The trained architecture and the easy-to-access performance of each architecture might provide some insidious ways for designing algorithms to over-fit the best architecture in our {\NAME}. Thus, we propose some rules to follow in order to achieve the original intention of {\NAME}, a fair and efficient benchmark.

\textit{1. No regularization for a specific operation.} Since the best architecture is known in our benchmark, specific designs to fit the structural attributes of the best performing architecture constitute one of the insidious ways to fit our {\NAME}.
For example, as mentioned in~\Secref{sec:benchmark}, we found that the best architecture with the same number of parameters for CIFAR-10 on {\NAME} is ResNet. Restrictions on the number of residual connections is a way to over-fit the CIFAR-10 benchmark.
While this can give a good result on this benchmark, the searching algorithm might not generalize to other benchmarks.

\textit{2. Use the same meta hyperparameter for different datasets and search spaces in {\NAME}.
}
The searching algorithm has some meta hyperparameter that controls the behaviour of search. For example, the temperature $\tau$ in GDAS or the band width factor in BOHB.
Using the same meta hyperparameter could evaluate the robustness of the searching algorithm and prevent it from over-fitting to a specific dataset.\looseness=-1

\textit{3. Use the provided performance.}
The training strategy affects the performance of the architecture. We suggest to stick to the performance provided in our benchmark even if it is feasible to use other $\gH$ to get a better performance. This provides a fair comparison with other algorithms.

\textit{4. Report results of multiple searching runs.} Since our benchmark can help to largely decrease the computational cost for a number of algorithms, multiple searching runs, which give stable results of the searching algorithm with acceptable time cost, are strongly recommended.

\textbf{Limitation with regard to hyperparameter optimization (HPO).}
The performance of an architecture depends on the hyperparameters $\gH$ for its training and the optimal configuration of $\gH$ may vary for different architectures. In {\NAME}, we use the same configuration for all architectures, which may bring biases to the performance of some architectures. 
One related solution is HPO, which aims to search for the optimal hyperparameter configuration. However, searching for the optimal hyperparameter configurations and the architecture in one shot is too computationally expensive and still is an open problem~\cite{dong2020autohas}.

\textbf{Potential extension of {\NAME}.}
Despite the straightforward extension by introducing HPO into {\NAME}, there are some other interesting directions.
One tendency in NAS is the cost constrained searching. For example, how to design a FLOPs constrain loss to regularize the discovered architecture to be efficient~\cite{wan2020fbnetv2,cai2020once,dong2019network}? Since the latency and FLOPs information are off-the-shelf in {\NAME}, our {\NAME} can also be used to benchmark NAS algorithms using different kinds of cost loss.

\textbf{Potential designs using diagnostic information in {\NAME}}.
As pointed in \Secref{sec:nas-diagnostic-info}, different kinds of diagnostic information are provided.
We hope that more insights about NAS could be found by analyzing these diagnostic information and further motivate potential solutions for NAS.
For example, parameter sharing~\cite{pham2018efficient} is the crucial technique to improve searching efficiency, but shared parameter would sacrifice the accuracy of each architecture.
Could we find a better way to share parameters of each architecture from the learned thousands of models' parameters?
Could we design new algorithms to take the mutual benefits of both multi-trial and weight sharing based methods?

\textbf{Generalization ability of the search space}.
It is important to test the generalization capability of the empirical observations on this dataset.
One possible strategy is to do all benchmark experiments on a much larger search space.
Unfortunately, it is prohibitive regarding the expensive computational cost.
We bring some results from \cite{ying2019bench,arber2020nas1shot1,peng2020pyglove} to provide some preliminary evidence of generalization.
In \Figref{fig:fed-nas-alg-results}, we show the rankings of RANDOM, REA, and REINFORCE is (REA $\geq$ REINFORCE $\geq$ RANDOM).
This is consistent with results in NAS-Bench-101, which contains more architecture candidates.
For NAS methods with parameter sharing, we find that GDAS $\geq$ DARTS (2nd) $\geq$ DARTS (1st), which is also consistent with results in NAS-Bench-1SHOT1.
Therefore, though it is not guaranteed, observations from our {\NAME} have a potential to generalize to other search spaces.

\textbf{Acknowledgments:}~We would like to thank Gabriel Bender, Pieter-Jan Kindermans, and Hanxiao Liu for their suggestions on experiments.
Part of this project was supported by Google Cloud Credits from GCP Education Programs.

\bibliographystyle{IEEEtran}
\bibliography{IEEEabrv,ms}

\appendix

\section*{Application Programming Interface (API)}\label{sec:nas-301-api}

Users can easily query all information of an architecture by using our API, such as latency, training time, number of parameters, validation accuracy, etc.
In this section, we show some example codes to query them.

\begin{lstlisting}[language=Python, caption={Create an instance of our benchmark.}, backgroundcolor=\color{backcolour}, commentstyle=\color{codegreen}, keywordstyle=\color{magenta}, tabsize=2, stringstyle=\color{codepurple}, label={lst:create}, breaklines=true, basicstyle=\footnotesize]
from nats_bench import create
# Load the data of the topology search space
nats_bench = create(search_space='topology')
# Load the data of the size search space
nats_bench = create(search_space='size')
\end{lstlisting}

\begin{lstlisting}[language=Python, caption={Show the structure of each architecture.}, backgroundcolor=\color{backcolour}, commentstyle=\color{codegreen}, keywordstyle=\color{magenta}, tabsize=2, stringstyle=\color{codepurple}, label={lst:show}, breaklines=true, showspaces=false, keepspaces=true, showstringspaces=false, basicstyle=\footnotesize]
amount = len(nats_bench)
for i, arch_str in enumerate(nats_bench):
  print('{:}/{:} : {:}'.format(i, amount, arch_str))
\end{lstlisting}

\begin{lstlisting}[language=Python, caption={Query the data of 115$_{th}$ architecture when training with 90 epochs ($\gH^{1}$); find the architecture with the highest accuracy on the validation set of CIFAR-100.}, backgroundcolor=\color{backcolour}, commentstyle=\color{codegreen}, keywordstyle=\color{magenta}, tabsize=2, stringstyle=\color{codepurple}, label={lst:show_data_115}, breaklines=true, showspaces=false, keepspaces=true, showstringspaces=false, basicstyle=\footnotesize]
info = nats_bench.query_meta_info_by_index(
    arch_index=115, hp='90')
index, accuracy = nats_bench.find_best(
    dataset='cifar100', metric_on_set='valid')
\end{lstlisting}

\begin{lstlisting}[language=Python, caption={Query the configuration of 12$_{th}$ architecture, its cost information, and its performance on different datasets.}, backgroundcolor=\color{backcolour}, commentstyle=\color{codegreen}, keywordstyle=\color{magenta}, tabsize=2, stringstyle=\color{codepurple}, label={lst:show_info_12}, breaklines=true, showspaces=false, keepspaces=true, showstringspaces=false, basicstyle=\footnotesize]
config = nats_bench.get_net_config(
    arch_index=12, dataset='cifar10')
info = nats_bench.get_cost_info(
    arch_index=12, dataset='cifar10')
# The info is a dict, where key is train-loss,
# train-accuracy, train-all-time, test-loss, etc.
# The corresponding value is info[key].
info = nats_bench.get_more_info(
    arch_index=12, dataset='cifar10')
info = nats_bench.get_more_info(
    arch_index=12, dataset='cifar100')
info = nats_bench.get_more_info(
    arch_index=12, dataset='ImageNet16-120')
\end{lstlisting}

\begin{lstlisting}[language=Python, caption={More advanced features.}, backgroundcolor=\color{backcolour}, commentstyle=\color{codegreen}, keywordstyle=\color{magenta}, tabsize=2, stringstyle=\color{codepurple}, label={lst:show_advanced}, breaklines=true, showspaces=false, keepspaces=true, showstringspaces=false, basicstyle=\footnotesize]
# Query results of the 284-th architecture on
# CIFAR-100 when training with 12 epochs.
# The 'data' is a dict, where the key is the random
# seed and the value is the corresponding result.
data = nats_bench.query_by_index(
    arch_index=284, dataset='cifar100', hp='12')
# >> [777, 888, 999]
print(data.keys())
# Show the validation performance using the random
# seed of 888 for the 284-th architecture
info = results[888].get_eval('valid')
\end{lstlisting}

\section*{Comparison Under Different Training Epochs}

\begin{figure}[t!]
\centering
  \includegraphics[width=\linewidth]{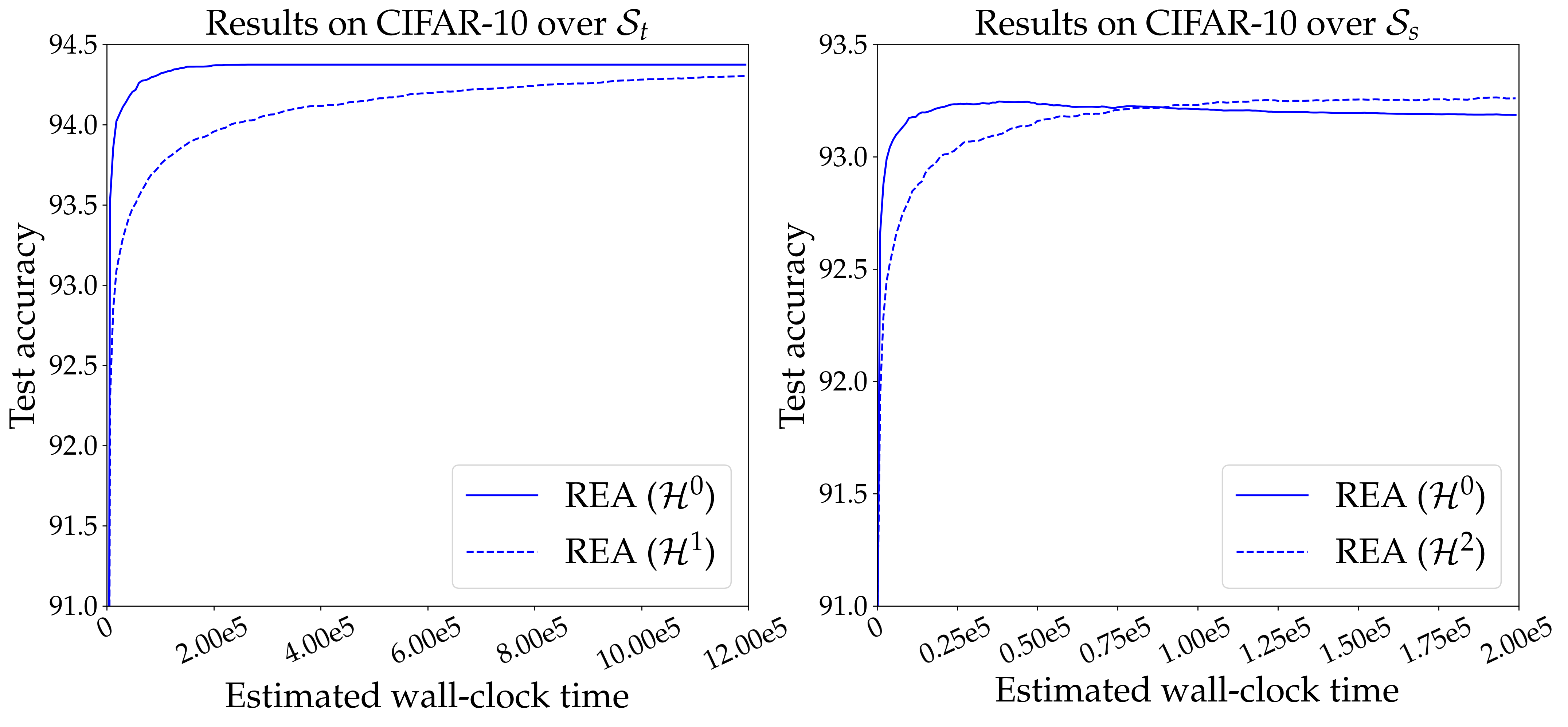}
\caption{
The test accuracy of REA discovered architectures on CIFAR-10 over two search spaces.
}
\label{fig:rea-h0-vs-h12}
\end{figure}

Multi-trial based algorithms are computationally expensive because they need to traverse many trials, and each trial may cost hours. On the contrary, low fidelity approximation is computationally efficient but provides less accurate feedback to train the model, such as using fewer training epochs for each trial in \Figref{fig:nas-alg-tss} and \Figref{fig:nas-alg-sss}.
To investigate the trade-off between efficiency and accuracy, we choose REA -- the best-performing multi-trial based algorithm on our NATS-Bench, and show its performance when using 12 epochs for training and 200 (or 90) epochs for training in \Figref{fig:rea-h0-vs-h12}.

For the topology search space, REA with $\gH^{0}$ (12 epochs) quickly converges after 1e5 seconds on CIFAR-10; however, REA with $\gH^{1}$ (200 epochs) is still worse than REA with $\gH^{0}$ after 1.2e6 seconds (about 333 hours).
For the size search space, REA with $\gH^{0}$ (12 epochs) quickly converges after 2e4 seconds.
REA with $\gH^{2}$ (90 epochs) is worse than REA with $\gH^{0}$ at the beginning, whereas it gradually outperforms REA with $\gH^{0}$ after about 8e4 seconds (about 22 hours). A similar phenomenon occurs on CIFAR-100 and ImageNet-16-120.

\begin{IEEEbiography}[{\includegraphics[width=1in,height=1.25in,clip,keepaspectratio]{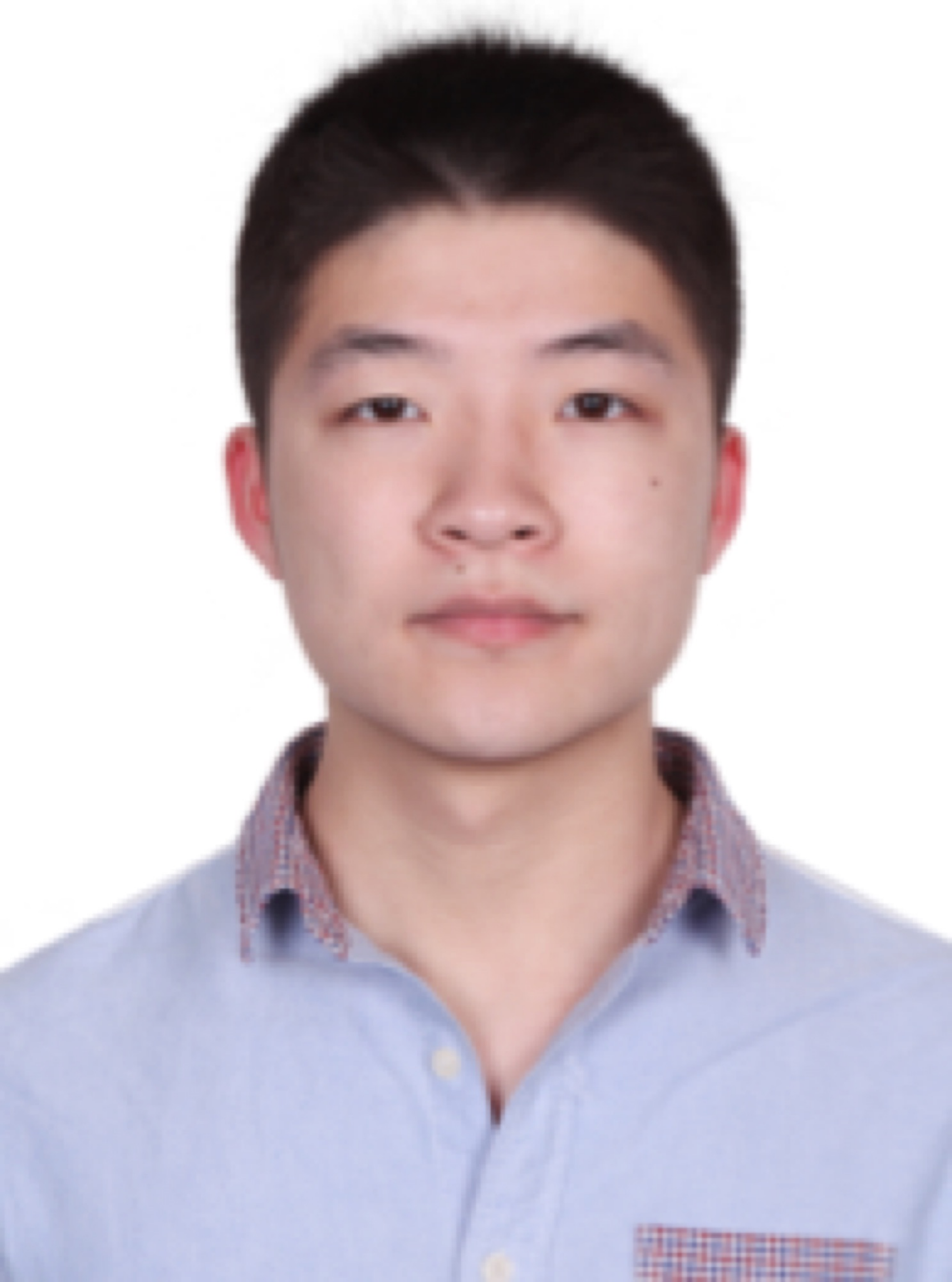}}]{Xuanyi Dong} received the B.E. degree in Computer Science and Technology from Beihang University (BUAA), Beijing, China, in 2016.
He is currently a Ph.D. student at School of Computer Science, University of Technology Sydney (UTS), Australia.
His research interests include automated deep learning and its application to real world problems.
In 2019, he received the Google PhD Fellowship.
\end{IEEEbiography}

\begin{IEEEbiography}[{\includegraphics[width=1in,height=1.25in,clip,keepaspectratio]{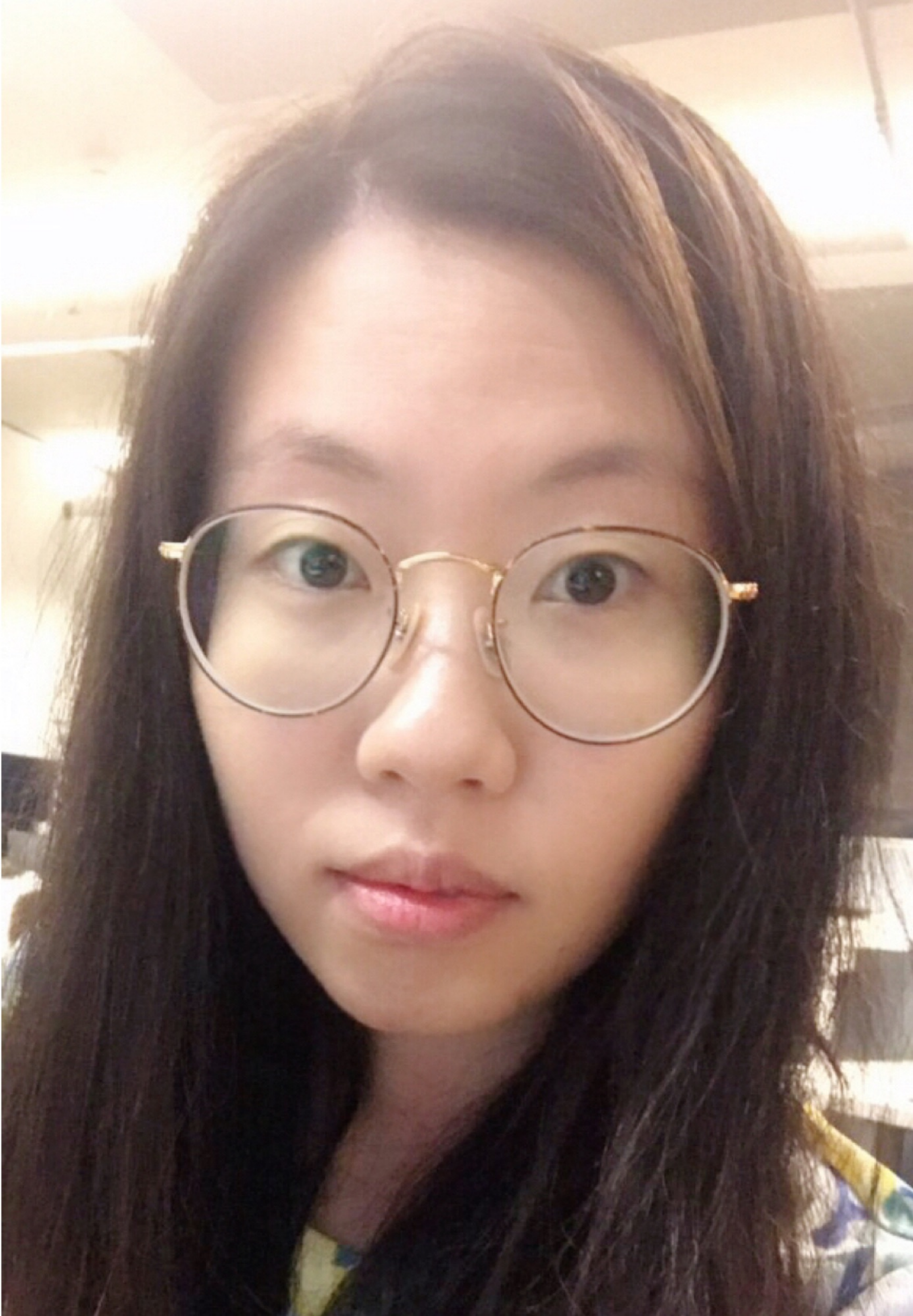}}]{Lu Liu}
received her bachelor’s degree from South China University of Technology (SCUT), Guangzhou, China, in 2017.
She is currently pursuing the Ph.D. at AAII, University of Technology Sydney (UTS), Australia.
Her current research interests include deep learning, machine learning, few-shot learning and meta-learning.
\end{IEEEbiography}

\begin{IEEEbiography}[{\includegraphics[width=1in,height=1.25in,clip,keepaspectratio]{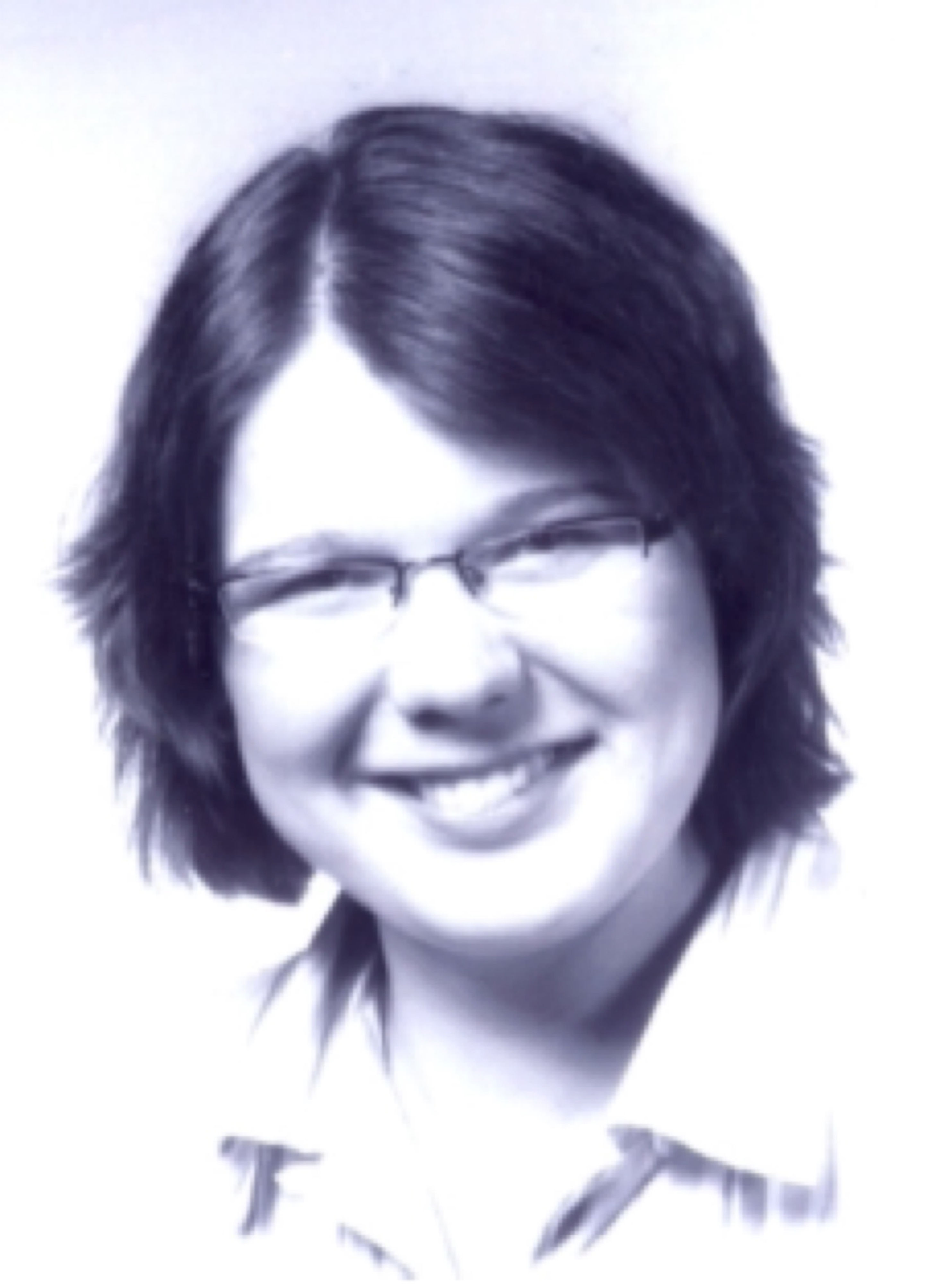}}]{Katarzyna Musial}
Katarzyna Musial received the M.Sc. degree in computer science from the Wrocław University of Science and Technology (WrUST), Poland, the M.Sc. degree in software engineering from the Blekinge Institute of Technology, Sweden, in 2006, and the Ph.D. from WrUST, in November 2009. In November 2009, she was appointed as a Senior Visiting Research Fellow with Bournemouth University (BU), where she has been a Lecturer in informatics, since 2010. In November 2011, she joined Kings as a Lecturer in computer science. In September 2015, she returned to Bournemouth University as a Principal Academic in Computing, where she was a member of the Data Science Institute. In September 2017, she joined as an Associate Professor in network science with the School of Software, University of Technology Sydney, where she is currently a member of the Advanced Analytics Institute. Her research interests include complex networked systems, analysis of their dynamics and its evolution, adaptive and predictive modeling of their structure and characteristics, as well as the adaptation mechanisms that exist within such systems are in the center of her research interests.
\end{IEEEbiography}

\begin{IEEEbiography}[{\includegraphics[width=1in,height=1.25in,clip]{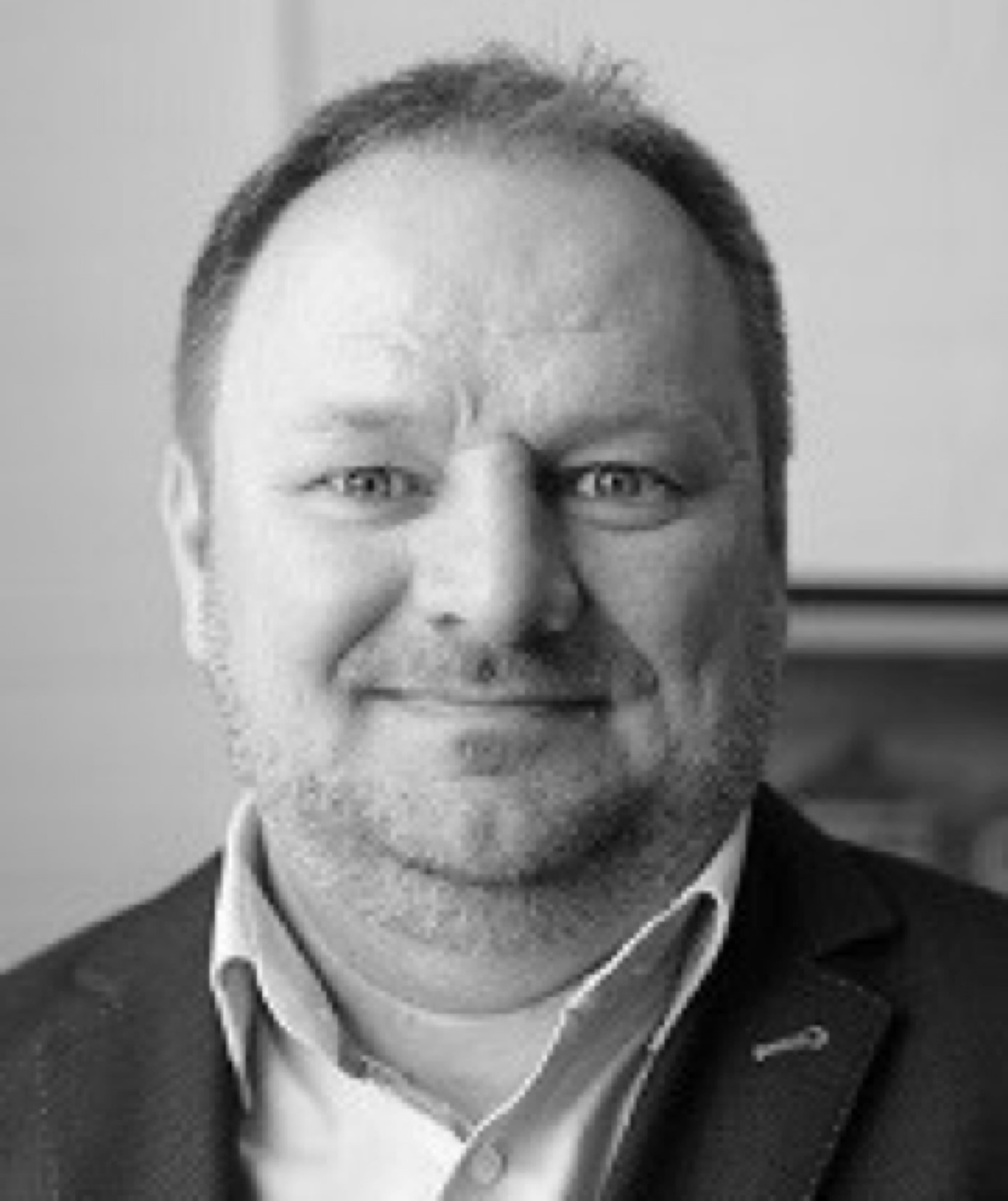}}]{Bogdan Gabrys}(SM'06) received the M.Sc. degree in electronics and telecommunication from Silesian Technical University, Gliwice, Poland, in 1994, and the Ph.D. degree in computer science from Nottingham Trent University, Nottingham, U.K., in 1998.
	
Over the last 25 years, he has been working at various universities and research and development departments of commercial institutions. He is currently a Professor of Data Science and a Director of the Advanced Analytics Institute at the University of Technology Sydney, Sydney, Australia. His research activities have concentrated on the areas of data science, complex adaptive systems, computational intelligence, machine learning, predictive analytics, and their diverse applications. He has published over 180 research papers, chaired conferences, workshops, and special sessions, and been on program committees of a large number of international conferences with the data science, computational intelligence, machine learning, and data mining themes. He is also a Senior Member of the Institute of Electrical and Electronics Engineers (IEEE), a Member of IEEE Computational Intelligence Society and a Fellow of the Higher Education Academy (HEA) in the UK. He is frequently invited to give keynote and plenary talks at international conferences and lectures at internationally leading research centres and commercial research labs. More details can be found at: http://bogdan-gabrys.com
\end{IEEEbiography}

\end{document}